\documentclass[lettersize,journal]{IEEEtran}
\usepackage{amsmath,amsfonts}
\usepackage{amssymb}
\usepackage{amsthm,bm}
\usepackage{algorithm}
\usepackage{algpseudocode}
\usepackage[font=small]{subcaption}
\usepackage{array}
\usepackage{textcomp}
\usepackage{multirow}
\usepackage{stfloats}
\usepackage{url}
\usepackage{verbatim}
\usepackage{graphicx}
\usepackage{cite}
\theoremstyle{plain}
\newtheorem{theorem}{Theorem}[section]

\newtheorem{assumption}[theorem]{Assumption}
\theoremstyle{definition}
\newtheorem{definition}[theorem]{Definition}

\theoremstyle{remark}

\def \E {\mathop{{\mathbb{E}}}\limits}

\def \N {\mathcal{N}}

\def \xb {\mathbf{x}}

\def \varepsilonb {\bm{\varepsilon}}

\def \taub {\bm{\taub}}

\def \E {\mathop{{\mathbb{E}}}\limits}

\def \N {\mathcal{N}}

\def \Xb {\mathbf{X}}
\def \Yb {\mathbf{Y}}
\def \Zb {\mathbf{Z}}

\def \Ab {\mathbf{A}}
\def \Bb {\mathbf{B}}

\def \Ib {\mathbf{I}}

\def \Mb {\mathbf{M}}

\def \Vb {\mathbf{V}}
\def \Wb {\mathbf{W}}

\def \oneb {\mathbf{1}}
\def \zerob {\mathbf{0}}

\begin{document}
\bibliographystyle{IEEEtran}
\title{Understanding Pan-Sharpening via Generalized Inverse}

\author{Shiqi Liu, Yihua Tan, Yutong Bai,  Alan Yuille
\thanks{Shiqi Liu and Yihua Tan are with Huazhong University of Science and Technology, Wuhan, 430074, Hubei, China(e-mail: shiqi.liu647@foxmail.com;yhtan@hust.edu.cn). }
\thanks{Yutong Bai is with University of California, Berkeley, Berkeley, 94720, California, United States(e-mail: ytongbai@gmail.com).}
\thanks{Alan Yuille is with Johns Hopkins University, Baltimore, 21218, Maryland,  United States.(e-mail: ayuille1@jhu.edu).}
\thanks{The prototype for this work was developed during an internship at the CCVL lab at Johns Hopkins University. It was later improved and finalized in Yihua Tan’s laboratory at Huazhong University of Science and Technology.}
\thanks{Yihua Tan is the corresponding author.}}



\markboth{Journal of \LaTeX\ Class Files,~Vol.~, No.~, July~2025}%
{Shell \MakeLowercase{\textit{et al.}}: A Sample Article Using IEEEtran.cls for IEEE Journals}


\maketitle

\begin{abstract}
Pan-sharpening algorithms utilize a panchromatic image and a multispectral image to generate a high spatial and high spectral image. However, the optimizations of the algorithms are designed with different standards. We employ a simple matrix equation to describe the Pan-sharpening problem. The conditions for the existence of a solution and the acquisition of spectral and spatial resolution are discussed. A down-sampling enhancement method is introduced to improve the estimation of spatial and spectral down-sample matrices. 

Using generalized inverse theory, we discovered two kinds of solution spaces of generalized inverse matrix formulations, which correspond to the two prominent classes of Pan-sharpening methods: component substitution and multi-resolution analysis. Specifically, the Gram-Schmidt adaptive method is demonstrated to align with the generalized inverse matrix formulation of component substitution. A model prior of the generalized inverse matrix of the spectral function is rendered. Theoretical errors are analyzed. The diffusion prior is naturally embedded with the help of general solution spaces of the generalized inverse form, enabling the acquisition of refined Pan-sharpening results.

 Extensive experiments, including comparative, synthetic, real-data ablation and diffusion-related tests are conducted. The proposed methods produce qualitatively sharper and superior results in both synthetic and real experiments.  The down-sampling enhancement method demonstrates quantitatively and qualitatively better outcomes in real-data experiments. The diffusion prior can significantly improve the performance of our methods across almost all evaluation measures. 
 
 The generalized inverse matrix theory helps deepen the understanding of Pan-sharpening mechanisms.
\end{abstract}

\begin{IEEEkeywords}
Pan-sharpening, Generalized Inverse, Image Fusion, Hyperspectral Processing, Matrix Theory, Diffusion Model.
\end{IEEEkeywords}

\section{Introduction}
Surpassing 70\% of optical observation satellites are equipped with low-resolution multispectral (MS) and high-resolution panchromatic (PAN) sensors \cite{zhang2012review}. Although high-resolution multispectral sensors can capture clear and colorful images, most optical observation satellites do not adopt this sensor. This situation is attributed to following scientific limitations.

\begin{itemize}
  \item Energy limitation\cite{zhang2012review}: The energy per unit area emitted into the optical sensor is limited, while the sensing capability of the sensor per unit is also restricted. The energy per unit area utilizes multiple optical sensors to receive; each sensor receives a portion of the energy. If several smaller sensors are tiled, each sensor receives the corresponding energy. This energy requires a sensor with a stronger sensing ability. Therefore, the energy emitted into the sensor must be negotiated with the sensing ability of the sensor.
  \item On-board storage and data transmission rate limitations\cite{zhang2012review}: The storage capacity of satellites is limited. Besides, the transmission velocity between the satellite and receiver platform is also limited. As a result, the use of high-resolution MS sensors means that higher transmission speeds are required.
\end{itemize}

To enhance vision performance using collected images from multispectral and panchromatic sensors, various image fusion methods have been proposed. Most of these methods can generally be classified into multi-resolution analysis and component substitution techniques\cite{vivone2014critical}.

Multi-Resolution Analysis (MRA) methods, also known as spatial methods, decompose the PAN image into multiple scales to extract its spatial information. Common decomposition techniques used in Pan-sharpening include Laplacian pyramids, wavelets, curvelets, and contourlets. MRA-based fusion offers advantages such as maintaining temporal coherence and spectral consistency, and is robust to aliasing, making it a promising area for future research.
In regard to multi-resolution analysis(MRA), there are following methods
\begin{itemize}
  \item High pass filtering methods \cite{chavez1991comparison}: Firstly, they extract the high-frequency part of the PAN image. Then, they consider the variance proportion of the valid image pixel value of the MS and PAN as the weight. Next, they add the weighted high-frequency part to the up-sampled multispectral image and finally achieve Pan-sharpening. While these methods represent an interesting trial, the measurement quantity of the high-frequency part of the PAN image is not transformed to the measurement quantity of the MS image. Consequently, the fusion performance, affected by the PAN image pixel scale, could be excessively blurred or sharp.
  \item Wavelet transformation methods\cite{vivone2013contrast}: They utilize wavelet transformation to extract different frequency components for fusion.
  \item MTF-GLP-HPM-R methods\cite{vivone2017regression}: They use a Generalized Laplacian Pyramid (GLP) with filters that match the Modulation Transfer Function (MTF) and high-pass modulation injection. This method also includes an initial step of spectral matching based on regression analysis.
  \item MTF-GLP-CBD methods\cite{aiazzi2006mtf}: They first extract the high-frequency component of the PAN image. Then, they use the quotient of the covariance between the low-resolution PAN and MS and the variance of low-resolution PAN to weight the high-frequency component. Finally, they add the weighted high-frequency component to the upsampled MS image to obtain the result.
  \item MTF-GLP-FS methods\cite{vivone2018full}: They address the estimation of full-resolution injection coefficients for regression-based Pan-sharpening.  Specifically, they focus on the effective GLP method using MTF-matched filter regression.  They propose and analyze an iterative algorithm that estimates these coefficients at full scale. 
      \end{itemize}

Component substitution (CS) methods, also known as spectral methods, enhance the spatial detail of MS images by projecting them into a different domain. This transformation separates the spatial information, which is then replaced with the high-resolution detail from a PAN image. Due to their straightforward implementation, CS techniques have been widely used in Pan-sharpening, with early examples including intensity-hue-saturation (IHS) and principal component analysis (PCA) methods from the 1990s. Regarding CS methods, there are following methods,
\begin{itemize}
  \item IHS methods\cite{carper1990use}: They first extract the intensity, hue, saturation bands from the MS image. Then, they utilize high resolution PAN image to replace the intensity band. Finally, they reconstruct the image. The fusion number of band is limited to three.
  \item PCA methods\cite{kwarteng1989extracting}: First, they perform principal component decomposition of a MS image. Then, they replace the first principal component with the PAN image. Finally, they apply an inverse transformation to reconstruct the high resolution image. However, these novel methods are limited by spectral and spatial inconsistencies, which may alter the spectral properties.
  \item Gram schmidt adaptive(GSA) methods\cite{laben2000process}, \cite{aiazzi2007improving}: They first  synthesize PAN by using the upsampled projected MS image. Then, they calculate the difference between the synthetic PAN and the original PAN. They use adaptive parameters to up-recover the difference value in different spectral channels. Finally, they add the up-recovered value to the spatial up-sampled multi-spectral image to obtain the high spatial and spectral resolution results.  
  \item C-GSA\cite{restaino2016context}: They determine the weights for injecting spatial detail from a PAN image into MS    channels by proposing a novel, context-adaptive (local) method for estimating these injection coefficients, based on image segmentation.  They assign region-specific coefficients, ensuring that all pixels within a segmented region share the same weights, while allowing for variations between regions.  A Binary Partition Tree segmentation, applied to the MS image, generates the regions used for coefficient calculation.
  \item BT-H\cite{lolli2017haze}: They use the optimized Brovey transform\cite{gillespie1987color}  and include haze correction\cite{lolli2017haze}.
  \item BDSD-PC\cite{vivone2019robust}: They extract spatial details that vary by band in Pan-sharpening while incorporating physical constraints.
\end{itemize}

Other methods, including variational optimization  methods and deep learning methods, have also been proposed to solve the Pan-sharpening problem.

Variational Optimization (VO) methods, which solve optimization models, have gained significant popularity due to advancements in convex optimization and inverse problems. These methods have been successfully applied in areas such as MS Pan-sharpening and hyperspectral (HS) image fusion.  VO approaches typically establish a model based on the relationship between the input PAN image, the low-resolution MS image, and the desired high-resolution MS image.  However, the underlying problem is inherently ill-posed, necessitating the use of regularizers that incorporate prior knowledge about the desired high-resolution MS image to constrain the solution space. The typical methods including SR-D \cite{vicinanza2014pansharpening} and TV\cite{palsson2013new}.

Deep learning (DL) techniques have become prevalent in  MS Pan-sharpening and HS image fusion. The first DL-based Pan-sharpening approach, \cite{huang2015new}, adapted an autoencoder inspired by sparse denoising.  In 2016,\cite{masi2016pansharpening} introduced the first fully convolutional neural network (CNN) for Pan-sharpening, termed the Pan-sharpening neural network (PNN), which used a three-layer architecture inspired by a super-resolution CNN.  Also in 2016, \cite{zhong2016remote} proposed a compressed sensing Pan-sharpening method using the Gram-Schmidt transform and a pre-existing super-resolution CNN.   These initial studies spurred significant research interest in DL-based Pan-sharpening, leading to numerous publications. Deep learning methods\cite{yang2023panflownet},\cite{cao2022proximal},\cite{fu2021model}, \cite{xie2019multispectral},\cite{wei2017boosting},\cite{yuan2018multiscale},\cite{he2019pansharpening},\cite{zhang2019pan},\cite{scarpa2018target},\cite{deng2020detail},\cite{yang2022memory},\cite{zhou2022pan},\cite{zhang2023spatial} incorporate more prior knowledge for Pan-sharpening tasks. Consequently, current research focuses on developing novel network architectures and pre-processing techniques to enhance generalization. 

There are numerous mathematical deduction of different CS and MRA methods\cite{vivone2014critical}\cite{meng2019review}. However, 
the implicit objectives of different CS and MRA methods were not fully optimized and still lacked exploration. What kind of mathematical models do CS and MRA methods solve, and what are their solution spaces? How to perform refined optimization in the solution space to obtain a better solution? These questions need to be explored and answered.

In this paper, based on the matrix equations of the Pan-sharpening problem, we discover two kinds of solution spaces regarding the generalized inverse solution expression. According to the generalized inverse representation, we find that theses two solution spaces correspond to the general CS class of methods and the general MRA class of methods. Choosing different generalized inverse matrices yields different methods. Furthermore, it yields down-sampling enhancement for better fusion results. We show that the GSA method corresponds precisely to the generalized inverse expression of CS methods. We then embed prior knowledge in the range of the generalized inverse of the spectral response matrix. We analyze the errors in different cases. Additionally, we demonstrate that if the down-sampling enhancement is implemented, then CS methods and MRA methods produce the same results. To further incorporate the prior knowledge of the diffusion model, we substitute the general solution spaces of the generalized inverse form into the evidence lower bound of the diffusion model and optimize the auxiliary matrix to obtain a refined Pan-sharpening solution. Based on the matrix representation, we propose five different evaluation indicators. Both synthetic and real data experiments show that the results align with generalized inverse theory and that our method, incorporating prior and down-sampling enhancement, yields sharper qualitative results. The experiment conducted to validate the effect of the diffusion prior demonstrates that the diffusion prior can significantly improve the performance of our methods across almost all evaluation measures.

The main contributions are as follows:
\begin{itemize}
\item \textbf{The integration of generalized inverse theory with CS and MRA methods yields two general solution spaces, bridging the gap between mathematical theory and Pan-sharpening techniques.} This integration helps clarify the implicit objectives of different CS and MRA methods that were previously not fully explored. It provides a unified mathematical framework to understand how these two prominent classes of Pan-sharpening methods operate. By establishing this connection, it enables a deeper exploration of the solution spaces for further optimization.
\item \textbf{We derive the precise correspondence between GSA methods and the generalized inverse representation of CS methods.} This correspondence shows that GSA is a special case within the CS framework's generalized inverse representation, enhancing the understanding of GSA's mathematical foundation and its relationship with broader CS methods. It allows GSA to be further refined and optimized using potential prior knowledge, such as diffusion priors. 
\item \textbf{We improve the estimation of spatial and spectral response matrices using down-sampling enhancement}. Experiments demonstrate that this enhancement results in better quantitative and qualitative outcomes for both CS and MRA methods.
\item \textbf{Embedding prior knowledge in the range of the generalized inverse of the spectral response matrix helps obtain sharper reconstructed images.} Empirically, it is found that when elements of the generalized inverse are close to 1, the image is sharper, while smaller elements cause blurring. By restricting the generalized inverse's elements to a range around 1 (0.9-1.4), the method avoids blurred features. 
\item \textbf{Incorporating diffusion model prior knowledge into the generalized inverse solution spaces leads to refined Pan-sharpening results.} The general solution spaces of the generalized inverse form are substituted into the evidence lower bound of the diffusion model, and the auxiliary matrix is optimized using Score Distillation Sampling. Experiments validate that the diffusion prior significantly improves performance across almost all evaluation measures. This integration leverages the diffusion model's prior knowledge to further refine solutions within the general solution spaces. 
\end{itemize}


In the following sections, we denote the scalar, vector, matrix, and corresponding random variable in the
non-bold case, bold case, bold upper case, respectively.




\section{Methodology}

Suppose the desired high spatial resolution and high spectral resolution (HrHS) is $\Xb$, and we have high spatial resolution low spectral resolution (HrLS) $\Yb$ and low spatial resolution and high spectral resoltuion (LrHS) $\Zb$. We can express their relationship using the following model.
\subsection{Model}
\begin{assumption}\label{as:total equations}
$\Xb\in R^{HW\times S}$ is HrHS, $\Yb\in R^{HW\times s}$ is HrLS, $\Zb\in R^{hw\times S}$ is LrHS.
\begin{eqnarray}\label{eq:model}
\Yb = \Xb \Ab \\
\Zb = \Bb \Xb
\end{eqnarray}
where $\Ab \in R^{S\times s}$ is the spectral response function, $\Bb \in R^{hw\times HW}$ is the spatial response function, respectively.
\end{assumption}
$H$ is the height of high spatial resolution images, $W$ is the width of high spatial resolution images, $h$ is the height of low spatial resolution images,  $w$ is the width of low spatial resolution images, $S$ is the spectral number of high spectral images and $s$ is the spectral number of low spectral images. We mainly focus on the case where $s=1$. 

This model demonstrates that we can regard the HrLS $\Yb$ and  LrHS $\Zb$ as projections of HrHS $\Xb$ on different sides, determined by the  spectral response function $\Ab$ and the spatial response function $\Bb$, respectively. We also refer to $\Bb$ as the spatial down-sampling matrix and to $\Ab$ as the spectral down-sampling matrix.

\subsection{Definition of Generalized Inverse}
In order to introduce the solution for the model~\ref{as:total equations}, we first present the definition of the generalized inverse.
\begin{definition}\cite{ben2003generalized}
$\Ab^{-}$ is the generalized inverse of a matrix $\Ab$ if $\Ab\Ab^{-}\Ab=\Ab$. We define $\Ab\{-\}=\{\Ab^{-}|\Ab\Ab^{-}\Ab=\Ab\}$ to be the set of all the generalized inverse of $\Ab$.

Further, if $\Ab^{-}\Ab\Ab^{-}=\Ab^{-}$, $(\Ab^{-}\Ab)^T=\Ab^{-}\Ab$ and $(\Ab\Ab^{-})^T=\Ab\Ab^{-}$ then we define this $\Ab^{-}$ as the  Moore–Penrose inverse and denote it as $\Ab^{+}$.
\end{definition}

We can see that the generalized inverse inherits one property $\Ab\Ab^{-}\Ab=\Ab$ of the normal inverse.
\subsection{Properties}
\begin{theorem}\label{thm:solution exsits coditions}\cite{radhakrishna1967calculus}, the matrix equations in \textbf{Assumption}~\ref{as:total equations} have a solution if and only if \begin{eqnarray}
\Bb\Yb = \Zb\Ab\\
\Yb=\Yb\Ab^{-}\Ab\\
\Zb=\Bb\Bb^{-}\Zb.
\end{eqnarray}
where $\Ab^{-}$ refers to any generalized inverse of matrix $\Ab$, and $\Bb^{-}$ refers to any generalized inverse of matrix $\Bb$.
\end{theorem}

This theorem uses generalized inverses to establish the existence condition for the solution to the model.

%

%
%
%
%
\subsection{Solution for $\Ab$ and $\Bb$}


\begin{theorem}\label{thm:BY=ZA} We can use linear equations to rewrite the $\Bb\Yb-\Zb\Ab$ in Kronecker form
, i,e,
\begin{equation}
\begin{bmatrix}
\Zb_{hw\times S}\otimes\Ib_{s\times s},
-\Ib_{hw\times hw}\otimes\Yb^T_{s\times HW}
\end{bmatrix}
\begin{bmatrix}
vec(\Ab)\\
vec(\Bb)
\end{bmatrix}_{Ss+hwHW}
=
0,
\end{equation}
where the Kronecker product $\Zb\otimes\Ib =\begin{bmatrix}
                     z_{1,1}\Ib & z_{1,2}\Ib & \cdots & z_{1,S}\Ib \\
                     z_{2,1}\Ib &  z_{2,2}\Ib & \cdots & z_{2,S}\Ib\\
                     \vdots & \vdots &\ddots& \vdots \\
                     z_{hw,1}\Ib &  z_{hw,2}\Ib & \cdots & z_{hw,S}\Ib
                   \end{bmatrix}$  and $vec(\Ab)= [a_{11},\cdots,a_{S1},a_{12},\cdots,a_{S2},\cdots,a_{1s}\cdots,a_{Ss}]^T.$

 $\Bb\Yb=\Zb\Ab$ has a nonzero solution for $\Bb$ and $\Ab$\\ $\Leftrightarrow$ \begin{itemize}
                                                                                  \item ($HW\ge s$) or ($HW<s$ and $hw<\frac{Ss}{s-HW}$)
                                                                                  \item ($HW<s$ and $hw\ge\frac{Ss}{s-HW}$) and $rank(\begin{bmatrix}
\Zb_{hw\times S}\otimes\Ib_{s\times s},
-\Ib_{hw\times hw}\otimes\Yb^T_{s\times HW}
\end{bmatrix})<min(hws,Ss+hwHW)$.
                                                                                \end{itemize}
\end{theorem}
The subscripts on the matrices are just to emphasize the shape of each matrix. The theorem demonstrates that if the high-resolution space $HW$ is bigger than the low spectral number $s$, then there exists a non-zero solution for $\Ab$ and $\Bb$.

\textbf{Down-sampling enhancement:} By constructing down-sampling matrix $\hat{\Bb}$ to obtain $\Bb=\Zb\Zb^{+}\hat{\Bb}$, it yields $\Bb\Yb=\Zb(\Zb^{+}\hat{\Bb}\Yb)=\Zb\Ab$.  As a result, the solution turns out to be \begin{equation}
\Ab=\Zb^{+}\hat{\Bb}\Yb,
\end{equation} where $\Zb^{+}$ is the Moore-Penrose generalized inverse of $\Zb$. If we assume $rank(\Zb)=S$ it can be further represented as $\Zb^{+}=(\Zb^T\Zb)^{-1}\Zb^T$. 
The above formula also provides a optimal solution under the least square criterion \cite{planitz19793} for \begin{eqnarray}
\Ab=\arg\min\limits_{\Mb}\Vert \hat{\Bb}\Yb-\Zb\Mb\Vert.
\end{eqnarray}

%
In the subsequent theoretical discussion of error analysis, the use of down-sampling enhancement will induce the equivalence of the CS method and the MRA method. Experiments demonstrate that this enhancement results in better quantitative and qualitative outcomes for both CS and MRA methods.

\subsection{Solutions for $X$}
If we rewrite the form of \textbf{Assumption}~\ref{as:total equations} as linear equations with respect to $vec(\Xb)$, it yields
\begin{equation}
\begin{bmatrix}
\Ab^{T}\otimes\Ib\\
\Ib\otimes\Bb
\end{bmatrix}vec(\Xb)
=
\begin{bmatrix}
vec(\Yb)\\
vec(\Zb)
\end{bmatrix}.
\end{equation}

As we can see from the Kronecker form, the problem we want to solve is a system of linear equations. The following theorem will provide one kind of solution to the system of linear equations.
%

\begin{theorem}\label{thm:MRA theorem}\cite{radhakrishna1967calculus}, if the matrix equations in \textbf{Assumption}~\ref{as:total equations} have a solution, then one solution format to the \textbf{Assumption}~\ref{as:total equations} is the following:
\begin{eqnarray}
\Xb_{mra}=\Bb^{-}\Zb+(\Ib-\Bb^{-}\Bb)\Yb\Ab^{-},
\end{eqnarray}
According to \cite{ben2003generalized}, the general solution space is
\begin{eqnarray}
\Xb=\Xb_{mra}+(\Ib-\Bb^{-}\Bb)\Wb(\Ib-\Ab\Ab^{-}),
\end{eqnarray}
where $\Wb\in R^{HW\times S}$ is an arbitrary matrix.
\end{theorem}

This corresponds to the MRA method. The first term $\Bb^{-}\Zb$ corresponds to the up-sampling MS images. The second term $(\Ib-\Bb^{-}\Bb)\Yb\Ab^{-}$ corresponds to the spectral up-sampling of the difference between the PAN image and the low-pass PAN image. Different selections of $B^{-}$ and $A^{-}$ yield different MRA methods. Since we have derived the general solution space, we can use prior knowledge to further optimize $\Wb$, thereby refining the MRA methods.

Inspire by this solution, we derive another general solution, which is
\begin{eqnarray}
\Xb=\Bb^{-}\Zb + (\Yb-\Bb^{-}\Zb\Ab) \Ab^{-}.
\end{eqnarray}

\begin{theorem}\label{thm:CS theorem} If the matrix equations in \textbf{Assumption}~\ref{as:total equations} have a solution, then one solution format of the \textbf{Assumption}~\ref{as:total equations} is the following
\begin{eqnarray}\label{eq:CS equation}
\Xb_{cs}=\Bb^{-}\Zb + (\Yb-\Bb^{-}\Zb\Ab) \Ab^{-},
\end{eqnarray}
and the general solution space is
\begin{eqnarray}\label{eq:CS equation for all solution}
\Xb=\Xb_{cs}+(\Ib-\Bb^{-}\Bb)\Wb(\Ib-\Ab\Ab^{-}),
\end{eqnarray}
where $\Wb\in R^{HW\times S}$ is an arbitrary matrix.
\end{theorem}

This solution leads to the CS method. The first term $\Bb^{-}\Zb$ corresponds to the up-sampling MS images. The second term $(\Yb-\Bb^{-}\Zb\Ab) \Ab^{-}$ corresponds to the spectral up-sampling of the difference between the PAN image and the synthetic PAN image. Different selections of $B^{-}$ and $A^{-}$ yield different CS methods.  Since we have derived the general solution space for CS methods, we can use prior knowledge to further optimize $\Wb$, thereby refining the CS methods. The following theorem shows the relationship between two different representations.

\begin{theorem}\label{thm:CS=MRA theorem} If the matrix equation in \textbf{Assumption}~\ref{as:total equations} has a solution, then for fixed $\Ab^{-}\in \Ab\{-\}$ and $\Bb^{-}\in \Bb\{-\}$, $\Xb_{cs}=\Xb_{mra}$.
\end{theorem}

We will show that the traditional state-of-the-art method, GSA methods, can follow our CS generalized inverse solution representation.

\subsection{Gram-Schmidt adaptive method}
\begin{theorem}\label{thm:GSA}The Gram-Schmidt adaptive method can be described as follows. Suppose $var(\Zb\Ab)$ is invertible. Define
\begin{eqnarray}
\Wb =var(\Zb\Ab)^{-1} cov(\Zb\Ab,\Zb)\in \Ab\{-\}
\end{eqnarray}
where
\begin{eqnarray}
cov(\Zb\Ab,\Zb)&=&\frac{1}{hw}(\Zb\Ab-\oneb\frac{\oneb^T}{hw}\Zb\Ab)^T(\Zb-\oneb\frac{\oneb^T}{hw}\Zb),\nonumber
\end{eqnarray}
\begin{eqnarray}
var(\Zb\Ab) = \frac{1}{hw}(\Zb\Ab-\oneb\frac{\oneb^T}{hw}\Zb\Ab)^T(\Zb\Ab-\oneb\frac{\oneb^T}{hw}\Zb\Ab),
\end{eqnarray}
It yields
\begin{eqnarray}\label{eq:gsa equation}
\Xb_{gsa}=\Bb^{-}\Zb + (\Yb-\Bb^{-}\Zb\Ab) \Wb.
\end{eqnarray}

The Gram-Schmidt adaptive method can be viewed as a special case of the representation in Theorem~\ref{thm:CS theorem}. It yields
 \begin{eqnarray}
\Yb = \Xb_{gsa} \Ab \\
\Zb = \Bb \Xb_{gsa}.
\end{eqnarray}
\end{theorem}
Since GSA is a special case of the CS method, it shares the same representation of the general solution space. This allows it to be further refined and optimized using potential prior knowledge, such as a diffusion prior.
\subsection{Model prior}
\begin{figure*}[]
\centering
\captionsetup[subfigure]{font=tiny}
\begin{subfigure}{.22\textwidth}
	\centering
\includegraphics[width=\textwidth]{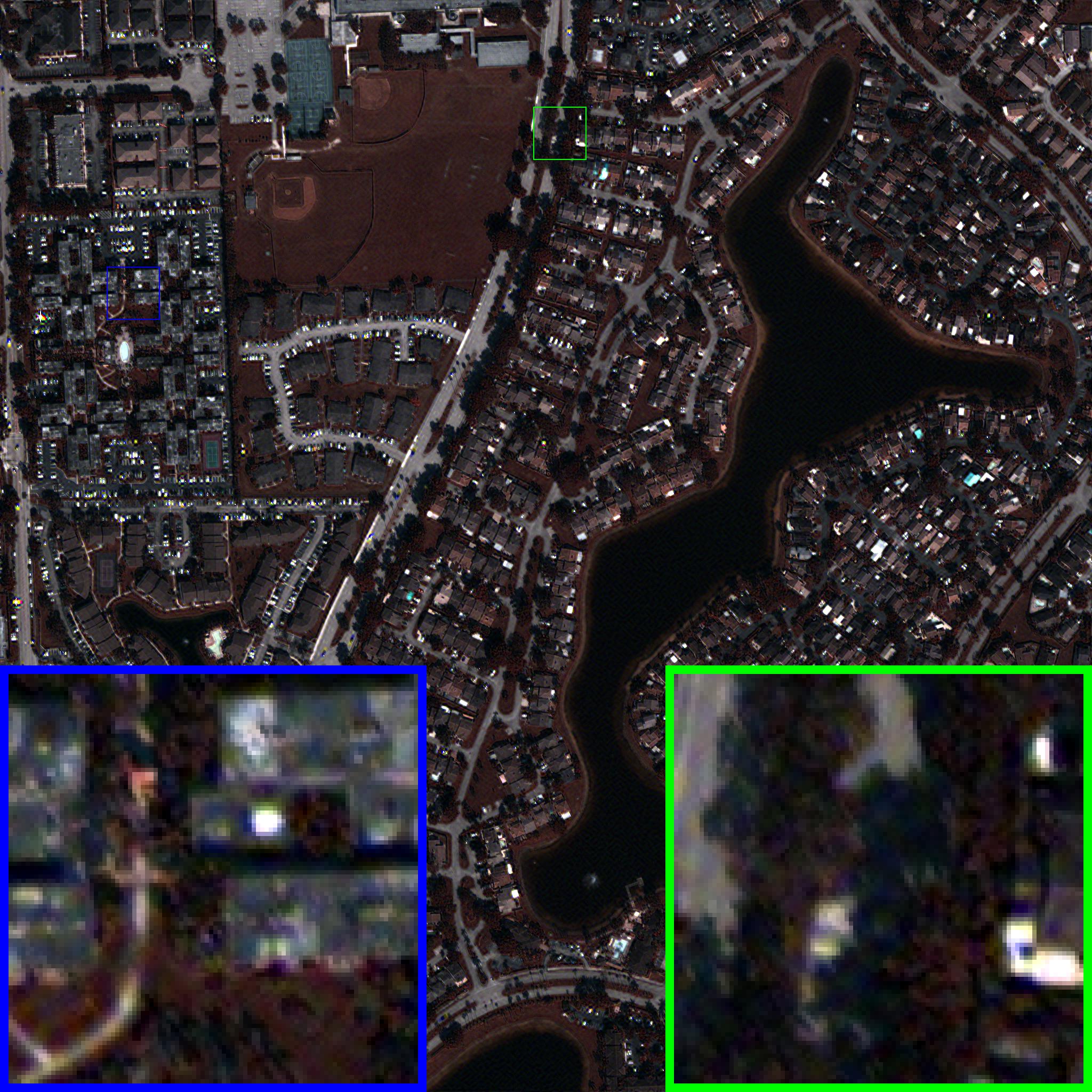}
	\caption{{GSA (band321)with weight ($0.84$,$0.45$,$0.29$)}}
\end{subfigure}
\hfill
\begin{subfigure}{.22\textwidth}
	\centering
\includegraphics[width=\textwidth]{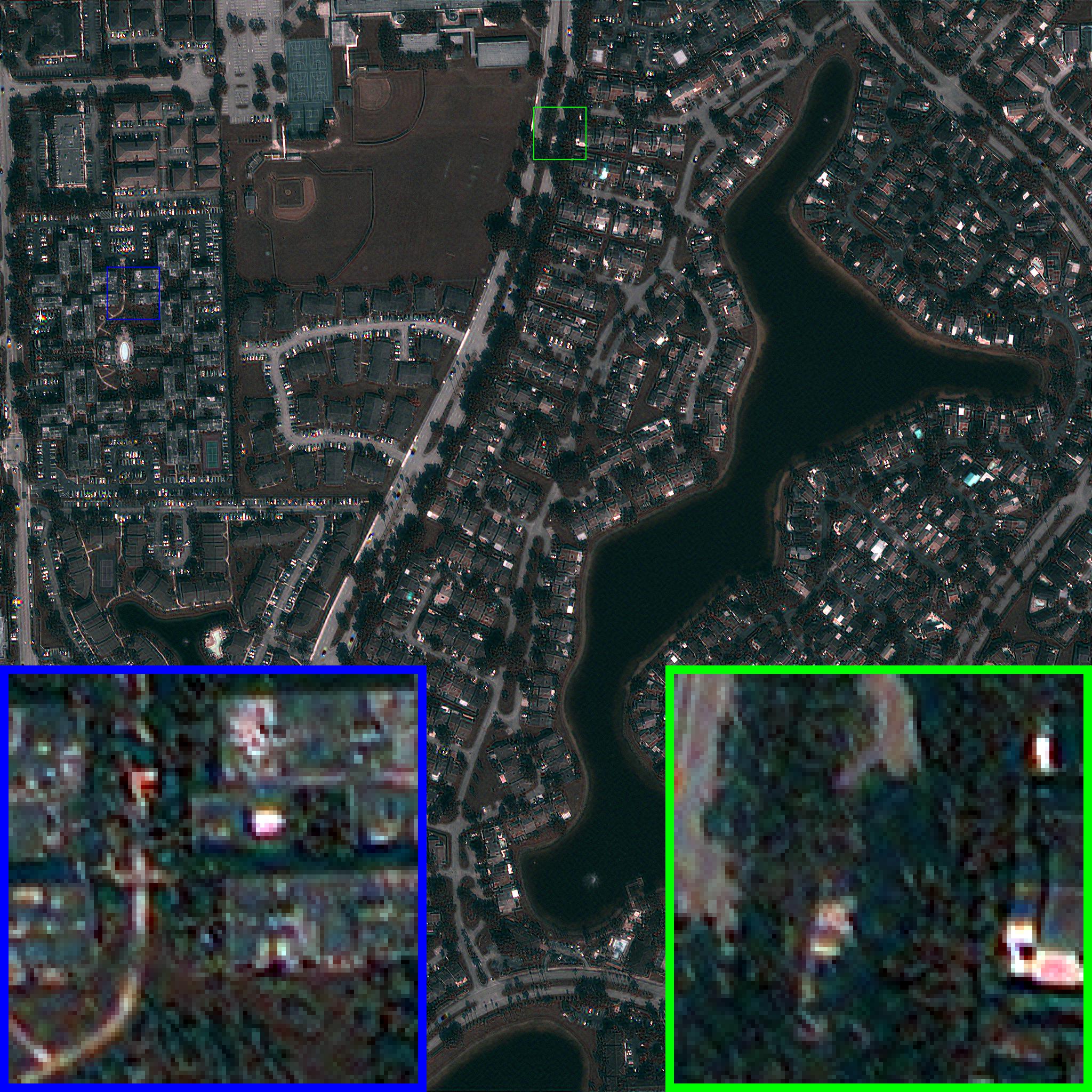}
	\caption{{PCS (band321) with weight ($0.90$,$1.40$,$0.90$)}}
\end{subfigure}
\hfill
\begin{subfigure}{.22\textwidth}
	\centering
\includegraphics[width=\textwidth]{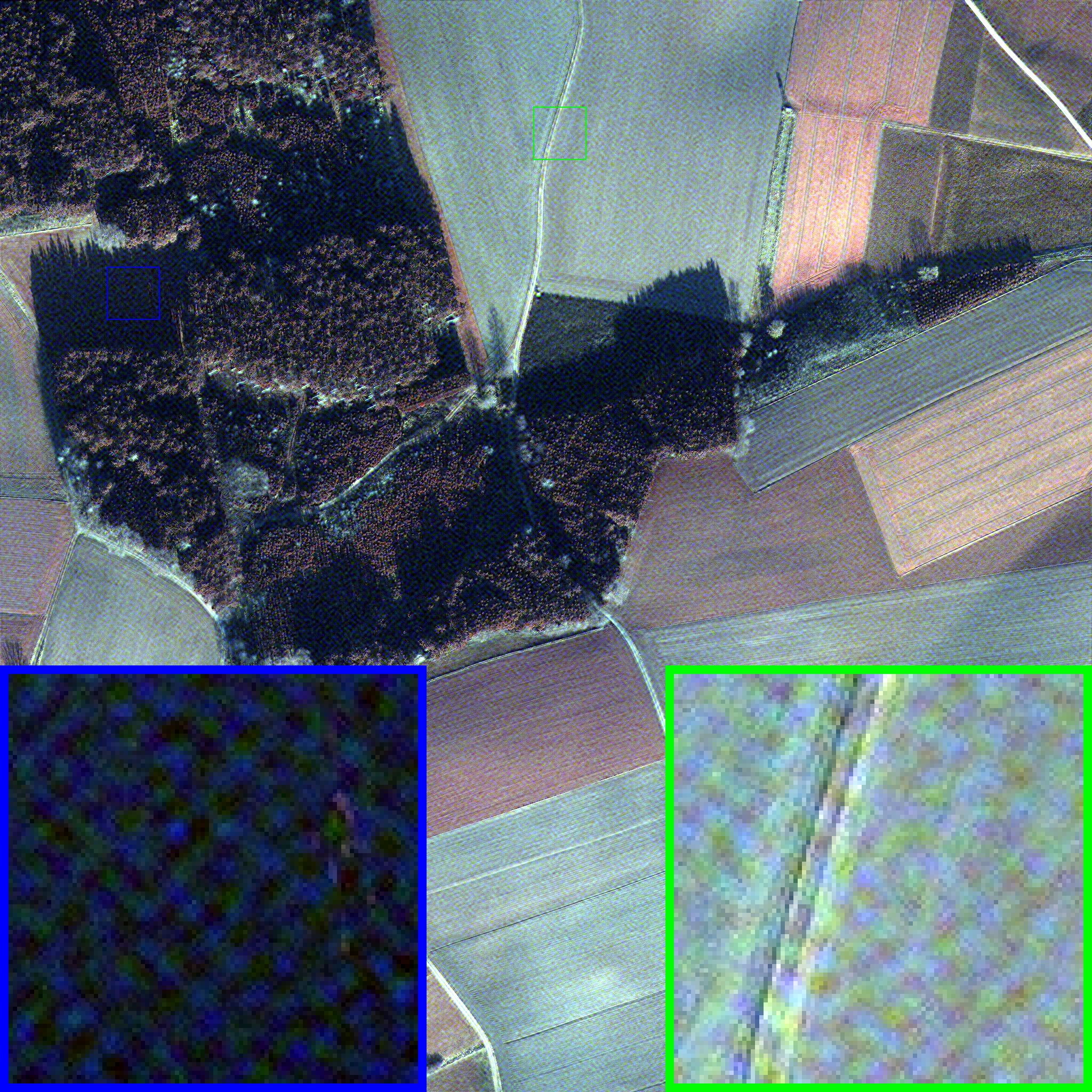}
	\caption{GSA (band321) with weight ($0.76$,$0.26$,$0.08$)}
\end{subfigure}
\hfill
\begin{subfigure}{.22\textwidth}
	\centering
\includegraphics[width=\textwidth]{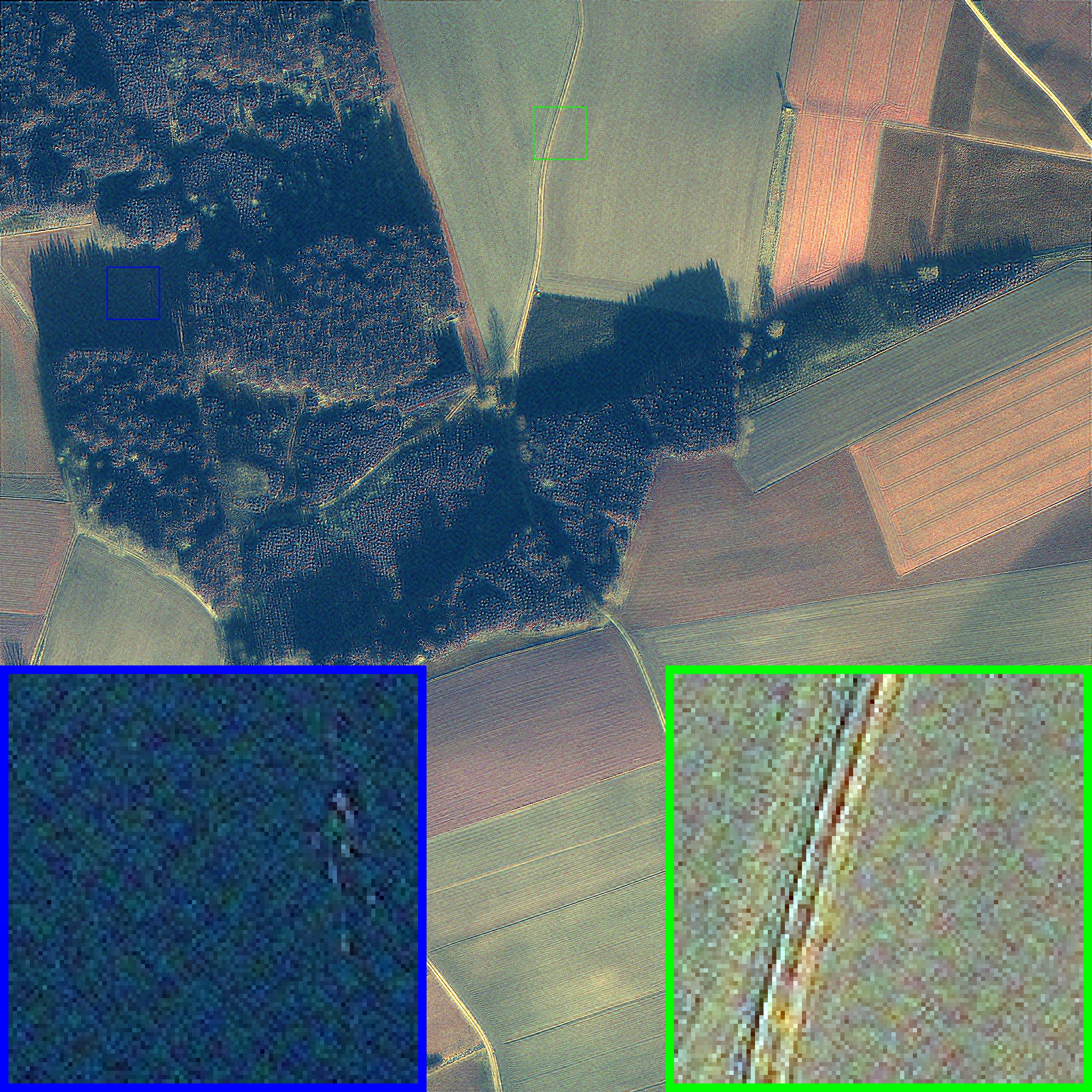}
	\caption{{PCS (band321) with weight ($0.90$,$0.90$,$0.90$)}}
\end{subfigure}
\caption{Visual comparisons of the synthetic results with different values of elements in $\Ab^{-}$\label{fig:compare element}. The first two images come from W2 Miam Mix dataset and second two images come from W3 Muni Nat dataset. We select the 1,2,3 band from the eight bands of the data to illustrate the difference. The differences between the images arise solely from the values of elements in $\Ab^{-}$. When the elements are close to 0, then the image becomes blurred. Conversely, when the elements are close to 1, then the image becomes sharp.}
\end{figure*}

An improper $\Ab^{-}$ will lead to unrealistic results for $\Xb$, especially causing blurred features. Empirically, we found that when the elements of $\Ab^{-}$ are  much smaller than 1, it  leads to a blurred image and when the elements of $\Ab^{-}$ are close to 1, it lead to a sharp image. The visual comparison of the differences induced by the elements values can be seen in the Figure~\ref{fig:compare element}. We can embed the prior knowledge of the high spatial and spectral resolution by empirically assuming the search space of ${A}^{-}$ to be the $Uniform([0.9,1.4]^{1\times S}\cap \{\Ab^{-}|\Ab\Ab^{-}\Ab=\Ab\})$\footnote{The paper's experiments mainly discuss the $s=1$ case} distribution among different sensors. 

If $s=1$, alternatively, it turns to be
\begin{eqnarray}
\Ab^{-}=\{\Mb\in [0.9,1.4]^{1\times S}|\Ab\Mb\Ab=\Ab\}.
\end{eqnarray}

%
%

\subsubsection{Algorithm}

The algorithms are presented in Algorithms~\ref{alg::PCS} and~\ref{alg::PMRA}. We calculate the computational complexity of our methods and the GSA method. The complexity for our two methods is $O(hwS\min(hw,S)+HWS+S^2)$. The complexity for the GSA method is $O(HWS+hwS\min(hw,S))$. The $S^2$ term can be omitted if $S$ is smaller than $HW$.
%
%
%
%
%

\begin{algorithm}[]
\caption{Prior Component substitution(CS) method}
\label{alg::PCS}
\begin{algorithmic}[1]
\Require
$(\Yb,\Zb,\Bb,\Bb^{-})$: The HrLS image, LrHS image, spatial response function matrix and up-sampling matrix;
\Ensure
$\Xb$
\State Construct the spatial up-sampling matrix $\Bb^{-}$ and spatial down-sampling response matrix  $\hat{\Bb}$ closely approximating  the sensor response function, satisfying $\hat{\Bb}\Bb^{-}=\Ib$.

\State Obtain $\Ab=\Zb^{+}\hat{\Bb}\Yb$ where $\Zb^{+}=(\Zb^T\Zb)^{-1}\Zb^T$.

\State Obtain $\Ab^{-}\in [0.9,1.4]^{1\times S}$ by solving $\Ab\Ab^{-}\Ab=\Ab$. It can be simplified to $\Ab^{-}\Ab=1$.

\State Synthesize low resolution low spectral LrLs $\Zb\Ab$.

\State Upsample $\Zb$ and $\Zb\Ab$ to obtain $\Bb^{-}\Zb$ and $\Bb^{-}\Zb\Ab$.

\State Obtain $\Xb=\Bb^{-}\Zb + (\Yb-\Bb^{-}\Zb\Ab) \Ab^{-}.$
\end{algorithmic}
\end{algorithm}

\begin{algorithm}[]
\caption{Prior Multiresolution analysis(MRA) method}
\label{alg::PMRA}
\begin{algorithmic}[1]
\Require
$(\Yb,\Zb,\Bb,\Bb^{-})$: The HrLS image, LrHS image, spatial response function matrix and up-sampling matrix;
\Ensure
$\Xb$
\State Construct the spatial up-sampling matrix $\Bb^{-}$ and spatial down-sampling response matrix  $\hat{\Bb}$ closely approximating to the sensor response function, satisfying $\hat{\Bb}\Bb^{-}=\Ib$.

\State Obtain $\Ab=\Zb^{+}\hat{\Bb}\Yb$ where $\Zb^{+}=(\Zb^T\Zb)^{-1}\Zb^T$.

\State Obtain $\Ab^{-}\in [0.9,1.4]^{1\times S}$ by solving $\Ab\Ab^{-}\Ab=\Ab$. It can be simplified to $\Ab^{-}\Ab=1$.

\State Synthesize low resolution low spectral(LrLs) image $\Bb\Yb$.

\State Upsample $\Zb$ and $\Bb\Yb$ to obtain $\Bb^{-}\Zb$ and $\Bb^{-}\Bb\Yb$

\State Obtain $\Xb=\Bb^{-}\Zb + (\Yb-\Bb^{-}\Bb\Yb) \Ab^{-}.$
\end{algorithmic}
\end{algorithm}

\subsection{Error analysis}

If we take the up-sampling operator $\Vb$ as $\Bb^{-}$, empirically $\Bb\Bb^{-} = \Ib$, but $\Bb\Vb = \Ib$ may not always hold. However, different algorithms may choose different $\Ab^{-}$. Negotiation with prior knowledge may lead to $\Ab^{-}$ not being in $\Ab\{-\}$. We denote $\Ab^{-}$ in this case as $\Wb$. Further error analysis can be conducted.
\subsubsection{Component substitution form analysis}
\begin{theorem}\label{thm:error XA for cs}If $\Xb_{recover}=\Vb\Zb+(\Yb-\Vb\Zb\Ab)\Wb$, then $\Xb_{recover}\Ab -\Yb = (\Vb\Zb\Ab - \Yb)(\Ib-\Wb\Ab).$
\end{theorem}

\begin{theorem}\label{thm:error BX for cs}If $\Xb_{recover}=\Vb\Zb+(\Yb-\Vb\Zb\Ab)\Wb$, then $\Bb\Xb_{recover} -\Zb = (\Bb\Vb-\Ib)\Zb+(\Bb\Yb-\Bb\Vb\Zb\Ab)\Wb$.
\end{theorem}

These two theorems demonstrate the spatial error and spectral error induced by up-sampling operator and $\Ab^{-}$.

\subsubsection{Multiresolution analysis form analysis}

\begin{theorem}\label{thm:error XA for mra}If $\Xb_{recover}=\Vb\Zb+(\Yb-\Vb\Bb\Yb)\Wb$, then $\Xb_{recover}\Ab -\Yb = \Vb(\Zb\Ab-\Bb\Yb)-(\Ib-\Vb\Bb)\Yb(\Ib-\Wb\Ab)$.
\end{theorem}

\begin{theorem}\label{thm:error BX for mra}If $\Xb_{recover}=\Vb\Zb+(\Yb-\Vb\Bb\Yb)\Wb$, then $\Bb\Xb_{recover} -\Zb = (\Bb\Vb-\Ib)\Zb+(\Bb-\Bb\Vb\Bb)\Yb\Wb$.
\end{theorem}

These two theorems demonstrate the spatial error and spectral error induced by up-sampling operator and $\Ab^{-}$.

\begin{theorem}\label{thm: down sampling enhancement} With down-sampling enhancement(DSE), the prior component substitution method and prior multiresolution analysis methods are equivalent. That is 
If $\Bb=\Zb\Zb^{+}\hat{\Bb}$ and $\Ab=\Zb^{+}\hat{\Bb}\Yb$, then $\Xb_{mra}=\Vb\Zb+(\Ib-\Vb\Bb)\Yb\Ab^{-}=\Xb_{cs}=\Vb\Zb + (\Yb-\Vb\Zb\Ab) \Wb.$  
\end{theorem}
This theorem demonstrates that with down-sampling enhancement, the effect of our proposed methods can be the same. 

\subsubsection{Total Analysis}
\begin{theorem}\label{thm:total error}If the Assumption~\ref{as:total equations} has a solution, then the ground truth $\Xb$ and $\Xb_{mra}$ is different as follows,
\begin{eqnarray}
\Xb-\Xb_{mra}=(\Ib-\Bb^{-}\Bb)\Xb(\Ib-\Ab\Ab^{-}).
\end{eqnarray}
\end{theorem}
This theorem suggests that we could further determine $(\Ib-\Bb^{-}\Bb)\Xb(\Ib-\Ab\Ab^{-})$ to obtain a better solution.

\subsection{The diffusion prior for Pan-sharpening}
We could further employ diffusion prior to help determine best performance $\Xb$ as follows:
\begin{equation}\label{eqn:diffusion}
\max\limits_{\Wb} p_{diffusion\ model}(\Xb_{recovered}+(\Ib-\Bb^{-}\Bb)\Wb(\Ib-\Ab\Ab^{-})),
\end{equation}
where $\Xb_{recovered}$ can be $\Xb_{cs}$, $\Xb_{mra}$ or $\Xb_{gsa}$.
Since $p_{diffusion\ model}$ is intractable, we replace it with the evidence lower bound $ELBO_{\theta}$\cite{ho2020denoising} to derive  the following:
\begin{eqnarray}
  \max\limits_{\Wb} &&  ELBO_{\theta}(\Xb_{recovered}+(\Ib-\Bb^{-}\Bb)\Wb(\Ib-\Ab\Ab^{-}))\nonumber \\
   &=&  -\sum_{t=1}^{T}\E_{\varepsilonb_{0}} \Vert\hat{\varepsilonb}_{\theta}(\sqrt{\overline{\alpha}_t}(\Xb_{recovered}\nonumber\\ &&+(\Ib-\Bb^{-}\Bb)\Wb(\Ib-\Ab\Ab^{-})+\sqrt{1-\overline{\alpha}_t}\varepsilonb_{0},t)-\varepsilonb_0\Vert^2 \nonumber
\end{eqnarray}
where $\overline{\alpha}_t$ are the schedule parameters in the diffusion model\cite{ho2020denoising}, $\hat{\varepsilonb}_{\theta}$ is the U-Net function in the diffusion model\cite{ho2020denoising},  samples noises $\varepsilonb_0\sim \N(0,\Ib)$, $T$ is total time of the time scheduler of the diffusion model\cite{ho2020denoising}.

By sample $t$\cite{ho2020denoising}, it yields
\begin{eqnarray}
  \min\limits_{\Wb} &&     \E_{\varepsilonb_{0},t} \Vert\hat{\varepsilonb}_{\theta}(\sqrt{\overline{\alpha}_t}(\Xb_{recovered}+(\Ib-\Bb^{-}\Bb)\Wb(\Ib-\Ab\Ab^{-}))\nonumber \\ &&+\sqrt{1-\overline{\alpha}_t}\varepsilonb_{0},t)-\varepsilonb_0\Vert^2. \label{eqn:final objective}
\end{eqnarray}

One could use gradient descent on Equation~\ref{eqn:final objective} to optimize $\Wb$, thereby obtaining a refined result with diffusion prior by searching in the general solution space. We use Score Distillation Sampling\cite{poole2022dreamfusion} to accelerate optimization. Algorithm~\ref{alg:example} describes the each step of our model. It demonstrates that once we have a foundational diffusion model and identify the general solution spaces for CS and MRA methods with the help of generalized inverse theory, we can search within these solution spaces to refine Pan-sharpening techniques (for example, the GSA method).

\begin{algorithm}[tb]
   \caption{Pan-sharpening via Generalized-Inverse-Guided Diffusion Models}
   \label{alg:example}
\begin{algorithmic}
   \Require ($\Xb_{recovered}$, $\Ab$, $\Bb$, $\Ab^{-}$, $\Bb^{-}$, $\hat{\varepsilonb}_{\theta}$, $\bm{\overline{\alpha}}$, $m$) Pan-sharpened image, spectral down-sampling matrix, spatial down-sampling matrix, spectral up-sampling matrix, spatial up-sampling matrix, diffusion model, schedule, iteration steps  
   \Ensure 
   $\Xb_{recovered}+(\Ib-\Bb^{-}\Bb)\Wb(\Ib-\Ab\Ab^{-})$
   \State Initialize $\Wb$ as $\zerob$
   \For {$i=1$ {\bfseries to} $m$}
        \State Randomly initialize $\varepsilonb_{0}\sim \N(0,\Ib)$, $t\sim Uniform({1,\cdots,T})$.
        \State Take gradient step on  
   
        $\nabla_{\Wb} dot(stopgradient[\hat{\varepsilonb}_{\theta}(\sqrt{\overline{\alpha}_t}(\Xb_{recovered}+(\Ib-\Bb^{-}\Bb)\Wb(\Ib-\Ab\Ab^{-}))+ \sqrt{1-\overline{\alpha}_t}\varepsilonb_{0},t)-\varepsilonb_0],(\Xb_{recovered}+(\Ib-\Bb^{-}\Bb)\Wb(\Ib-\Ab\Ab^{-})))$
   \EndFor
\end{algorithmic}
\end{algorithm}

\section{Experiments and results}
We will first introduce the evaluation measures, and then the datasets that we use to compare our methods and finally provide the experiments on these datasets.
\subsection{Evaluation measures.} Five quantitative evaluation indices are employed for performance evaluation, including consistent root mean square error(Consistent RMSE), spatial root mean square error(Spatial RMSE),spectral root mean square error(Spectral RMSE), Inverse Ability, and root mean square error(RMSE).

\begin{definition}Consistent root mean square error(Consistent RMSE) refers to
\begin{equation}
\sqrt{\frac{1}{hws}\Vert\Zb\Ab-\Bb\Yb\Vert_F^2}.
\end{equation}
\end{definition}
The Consistent RMSE reflects the consistency of the spatial and spectral response function. Consistent RMSE$=0$ is a necessary condition for the problem to have a solution.

\begin{definition}Spatial root mean square error(Spatial RMSE) refers to
\begin{equation}
 \sqrt{\frac{1}{HWs}\Vert\Xb_{recovered}\Ab-\Yb\Vert_F^2}.
\end{equation}
The mean spatial root mean square error refers to the mean Spatial RMSE of the whole dataset.
\end{definition}
Spatial RMSE measures the spatial consistency between the recovered HrHS image and the HrLS image.

\begin{definition}Spectral root mean square error(Spectral RMSE) refers to
\begin{equation}
\sqrt{\frac{1}{hwS}\Vert \Bb\Xb_{recovered}-\Zb\Vert_F^2}.
\end{equation}
The mean spectral root mean square error refers to the mean Spectral RMSE of the entire dataset.
\end{definition}
Spectral RMSE measures the spectral consistency between the recovered HrHS image and the HrLS image.

\begin{definition}Inverse Ability refers to $\Ab^{-}\Ab$.
\end{definition}
The inverse ability of the $\Ab^{-}$ is an important index influencing both Spectral RMSE and Spatial RMSE. The closer $\Ab^{-}\Ab$ is to $1$, the better.

\begin{definition}Root mean square error(RMSE) refers to
\begin{equation}
\sqrt{\frac{1}{hws}\Vert\Xb-\Xb_{recover}\Vert_F^2}.
\end{equation}
\end{definition}
Root mean square error reflects the distance between the recovered image and the oracle image.

We also implemented SAM,EGRAS\cite{vivone2021benchmarking} for our synthetic experiments. However, we found they can not effectively reflect the advantage related to visual appearance(sharper results) of our method.  As a result, we didn't show them in our experiment tables. 

For the diffusion-related experiment, since ground truth data is available, we use RMSE, spatial RMSE, spectral RMSE, PSNR\cite{hore2010image}, SSIM\cite{hore2010image}, SAM, and ERGAS to evaluate our model’s performance and fully demonstrate its advantages.

\subsection{Dataset description.}
\textbf{Dataset description for the WV3 New York dataset\cite{deng2022machine}.}
The WorldView-3 ground sample distance is  0.31m for the PAN band and 1.24m for eight MS bands. WV3 imagery of New York City (USA) reveals a cityscape characterized by tall buildings. (The size of an MS spectral band is 512×512). Since many other algorithms have been compared on this dataset, we use this dataset to compare the spatial RMSE and spectral RMSE of our methods and the other methods, including CS methods, MRA methods, VO methods and Deep learning methods.

\textbf{Dataset description for the Chikusei dataset.} 
 The Chikusei dataset\cite{yokoya2016airborne}  consists of $2517 \times 2335$ pixels. The grid length is 2.5m. The central point is located at coordinates: 36.294946 degrees north, 140.008380 degrees east. The data was acquired at urban areas in Chikusei, Ibaraki, Japan on July 29, 2014. It has 128 bands. This dataset aims to verify the Pan-sharpening ability on HS data.
 
\textbf{Dataset description for the PAirMax dataset.} 
The PAirMax dataset\cite{vivone2021benchmarking} is composed of 14 PAN and MS image pairs which are collected over different landscapes by various satellites. The detailed parameters for different satellite images are as follows:
\begin{itemize}
  \item The GeoEye-1 ground sample distance is  0.46m for the PAN band and 1.84m for the blue, green, red and near-infrared bands.
  \item The WorldView-2 ground sample distance is  0.46m for the PAN band and 1.84m for eight MS bands.
  \item The WorldView-3 ground sample distance is  0.31m for the PAN band and 1.24m for eight MS bands.
  \item The WorldView-4 ground sample distance is  0.31m for the PAN band and 1.24m for the four MS bands.
  \item The SPOT-7 ground sample distance is  1.5m for the PAN band and  6m for the four MS bands.
  \item Pléiades-1B ground sample distance is  0.7m for the PAN band and 2.8m for the four MS bands.
\end{itemize}
The experiments on this dataset focus on ablation study.

\textbf{Dataset description for the PAirMax-RGB dataset.} 
To validate the performance of the diffusion model (trained on Imagenet\cite{dhariwal2021diffusion}, named ``256x256\_diffusion\_uncond.pt"), we created an RGB image sub-dataset derived from PAirMax. The RGB channels of ground truth image correspond to the RGB channels of the original MS\_LR image from the FF path. We cropped the upper-left $256\times256$ pixels region of the MS\_LR images and applied percentile stretching (0.25\%-99.75\%) to process the images, resulting in the final RGB ground truth images. For experiment simplification, the PAN image was generated by averaging the RGB image channels with equal weighting. This dataset contains a total of 9 ground truth images and is referred to as PAirMax-RGB.

\subsection{Comparison methods}
  We evaluate the different methods in comparison with state-of-art methods. 
  For WV3 New York dataset, the comparison methods include our methods(PCS and PMRA) and methods in the literature\cite{deng2022machine}\cite{vivone2020new}. Since we used the same dataset, same comparison methods and their codes\cite{deng2022machine}\cite{vivone2020new}, the visual appearances of different comparison methods are consistent with the results in \cite{deng2022machine}.   

  For Chikusei dataset and  PAirMax dataset, we focus on the comparison of our methods with those methods that align with the generalized inverse illustration. The comparison methods include: GSA\cite{wang2005comparative}, MTF-GLP-CBD\cite{alparone2007comparison}, Prior component substitution(PCS), Prior multiresolution analysis (PMRA).For PAirMax dataset, we also compare the methods with(without) the down-sampling enhancement(DSE).
  
  For PAirMax-RGB, we focus on the validating the improvement of our methods combing with the diffusion priors. The comparison methods include: GSA, GSA+diffusion, MTF-GLP-CBD, PCS, and PCS +diffusion.

\subsection{ Experimental model training settings}
For the diffusion model used in Pan-sharpening, we implemented the experiment using PyTorch. The training was conducted on an NVIDIA GeForce RTX 3060(12G). The diffusion process employs a linear schedule with 1,000 timesteps. We utilize the Adam optimizer\cite{kingma2014adam} with a learning rate of $10^{-2}$ for gradient-based optimization of the parameter matrix $\Wb$. Approximately $2000$ iterations are required to optimize $\Wb$, and the process takes roughly 10 minutes to produce a single image. 

\subsection{Comparative experiments: performance comparison on WV3 New York}
In the experiment, bilinear down-sampling was selected as the spatial down-sampling matrix for calculating the spectral RMSE. The spectral weighting matirx was computed using our proposed method. Here we mainly present the full-resolution experiments to highlight the good consistency of our method.
\subsubsection{Quantitative results}
According to table~\ref{tab:Comparison experiments on WV3 New York Dataset}, the quantitative result demonstrates that PCS/PMRA methods(with down-sampling enhancement) are relatively better than the other 16 methods except for the spectral RMSE of the EXP method. Since EXP method is the interpolation of MS image using a polynomial kernel, its spatial RMSE is notably high. In contrast, both the spatial RMSE and spectral RMSE vaules for the  PCS/PMRA Methods (with down-sampling enhancement) are near zero. 
\begin{table}[]
\centering
\caption{Full-resolution comparison experiments on WV3 New York Dataset}
\label{tab:Comparison experiments on WV3 New York Dataset}
\resizebox{0.48\textwidth}{!}{
\begin{tabular}{lcc}
\hline
\textbf{Method} &\textbf{Spectral RMSE}($\downarrow$)&\textbf{Spatial RMSE}($\downarrow$)\\\hline
EXP\cite{deng2022machine}&\textbf{4.5452}&109.2157\\\hline
BT-H\cite{lolli2017haze}&71.6169&\underline{17.8711}\\\hline
BDSD-PC\cite{vivone2019robust}& 114.9423&74.8058\\\hline
C-GSA\cite{restaino2016context}&75.4264&26.8394\\\hline
SR-D\cite{vicinanza2014pansharpening}&100.4333&107.7014\\\hline
MTF-GLP-HPM-R\cite{vivone2017regression}&74.2957&63.7501\\\hline
MTF-GLP-FS\cite{vivone2018full}&71.2183&58.1442\\\hline
TV\cite{palsson2013new}&97.7862&61.4820\\\hline
PanNet\cite{yang2017pannet}&83.1296&71.4575\\\hline
DRPNN\cite{wei2017boosting}&83.2450&42.0010\\\hline
MSDCNN\cite{yuan2018multiscale}&84.7197&45.3580\\\hline
BDPN\cite{zhang2019pan}&85.6572&39.7455\\\hline
DiCNN\cite{he2019pansharpening}&88.2270&73.7617\\\hline
PNN\cite{masi2016pansharpening}&121.9181&69.4336\\\hline
APNN\cite{scarpa2018target}&79.4560&52.2568\\\hline
FusionNet\cite{deng2020detail}&107.6745&122.9544\\\hline
\textbf{PCS/PMRA} &\underline{10.0245}& \textbf{5.4193}\\\hline
\end{tabular}}
\end{table}
\subsubsection{Qualitative results}
\begin{figure*}[]
\centering
\captionsetup[subfigure]{font=tiny}
\begin{subfigure}[t]{.24\textwidth}
	\centering
\includegraphics[width=\textwidth]{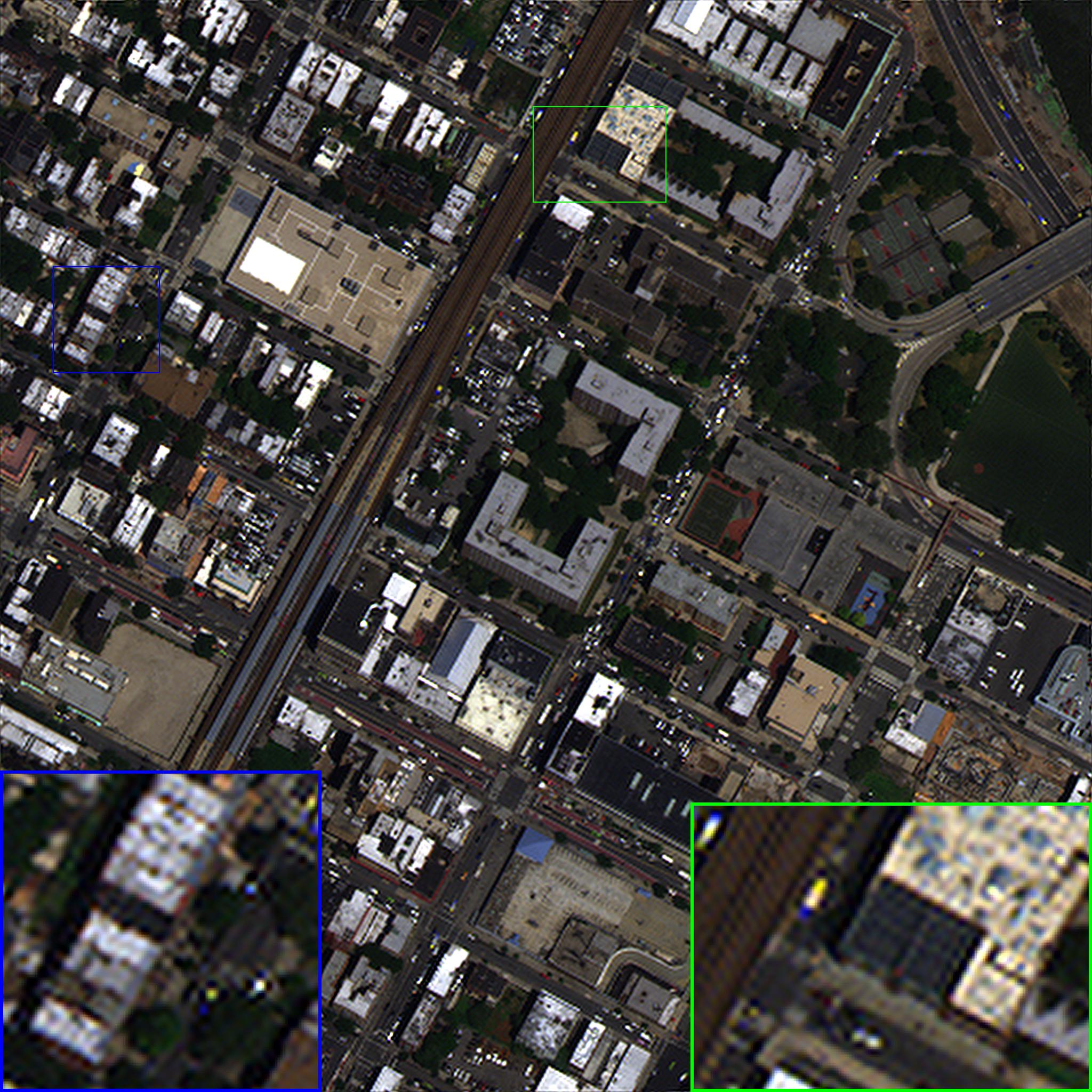}
	\subcaption{{EXP}}
\end{subfigure}
\hfill
\begin{subfigure}[t]{.24\textwidth}
	\centering
\includegraphics[width=\textwidth]{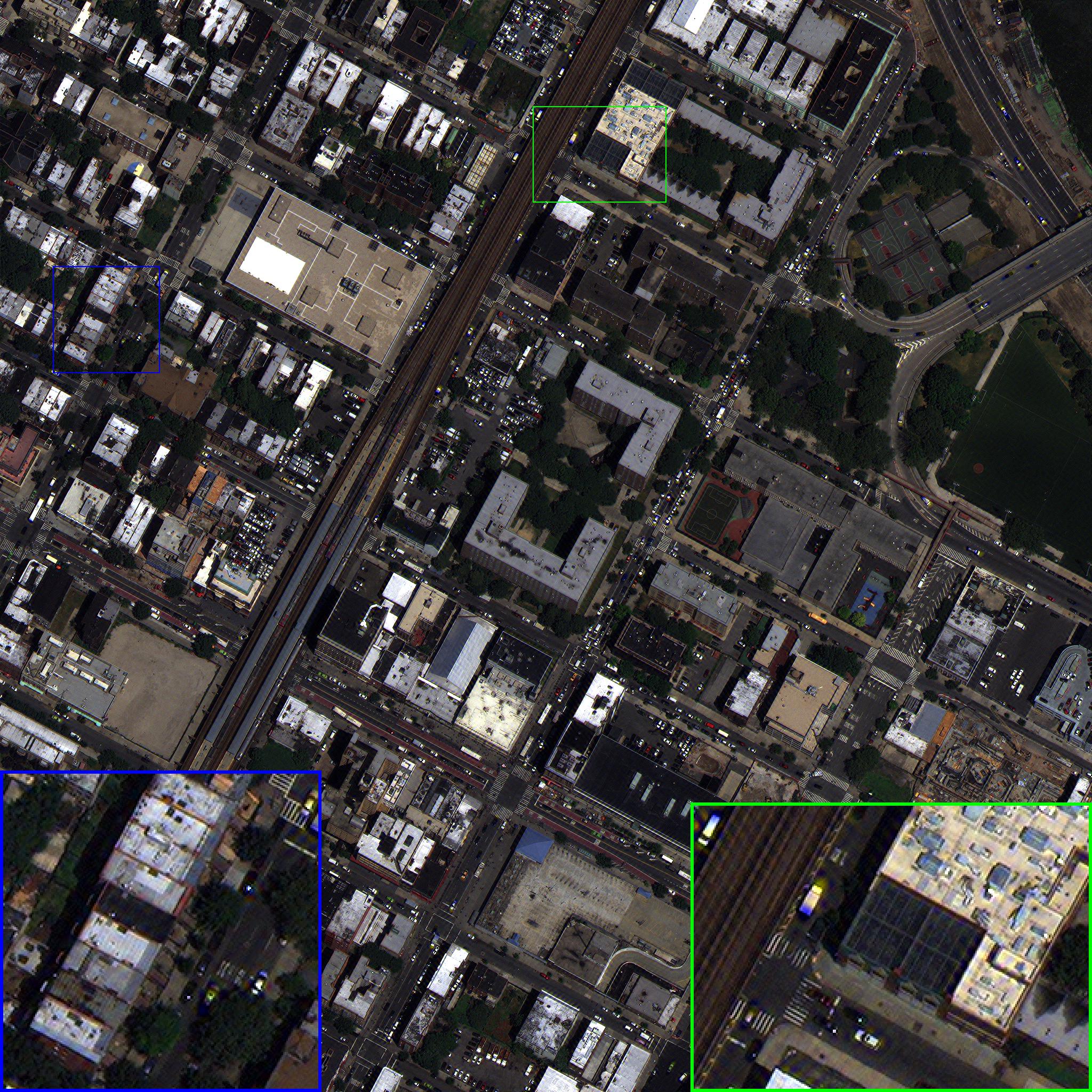}
	\subcaption{{BT-H}}
\end{subfigure}
\hfill
\begin{subfigure}[t]{.24\textwidth}
	\centering
\includegraphics[width=\textwidth]{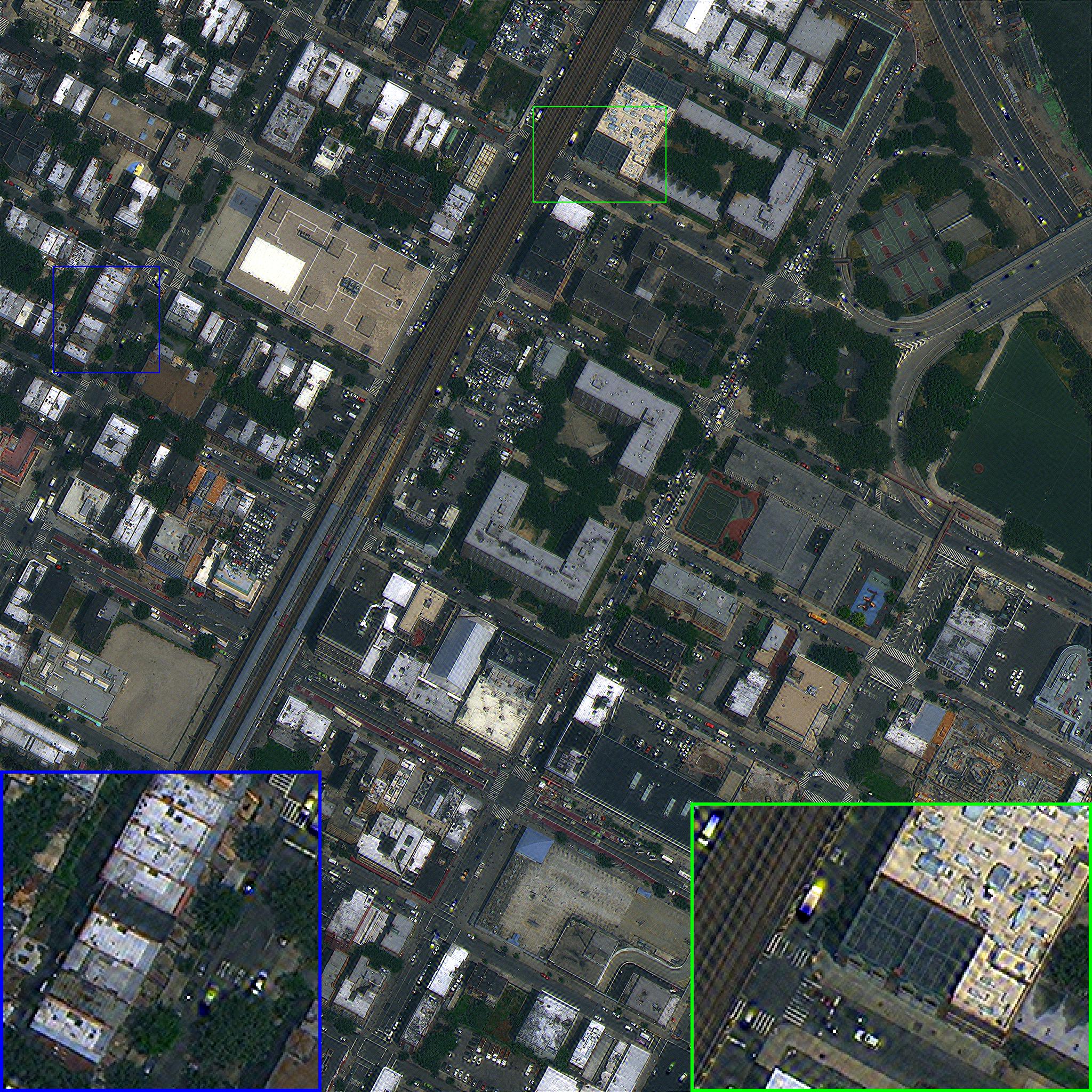}
	\subcaption{{BDSD-PC}}
\end{subfigure}
\hfill
\begin{subfigure}[t]{.24\textwidth}
	\centering
\includegraphics[width=\textwidth]{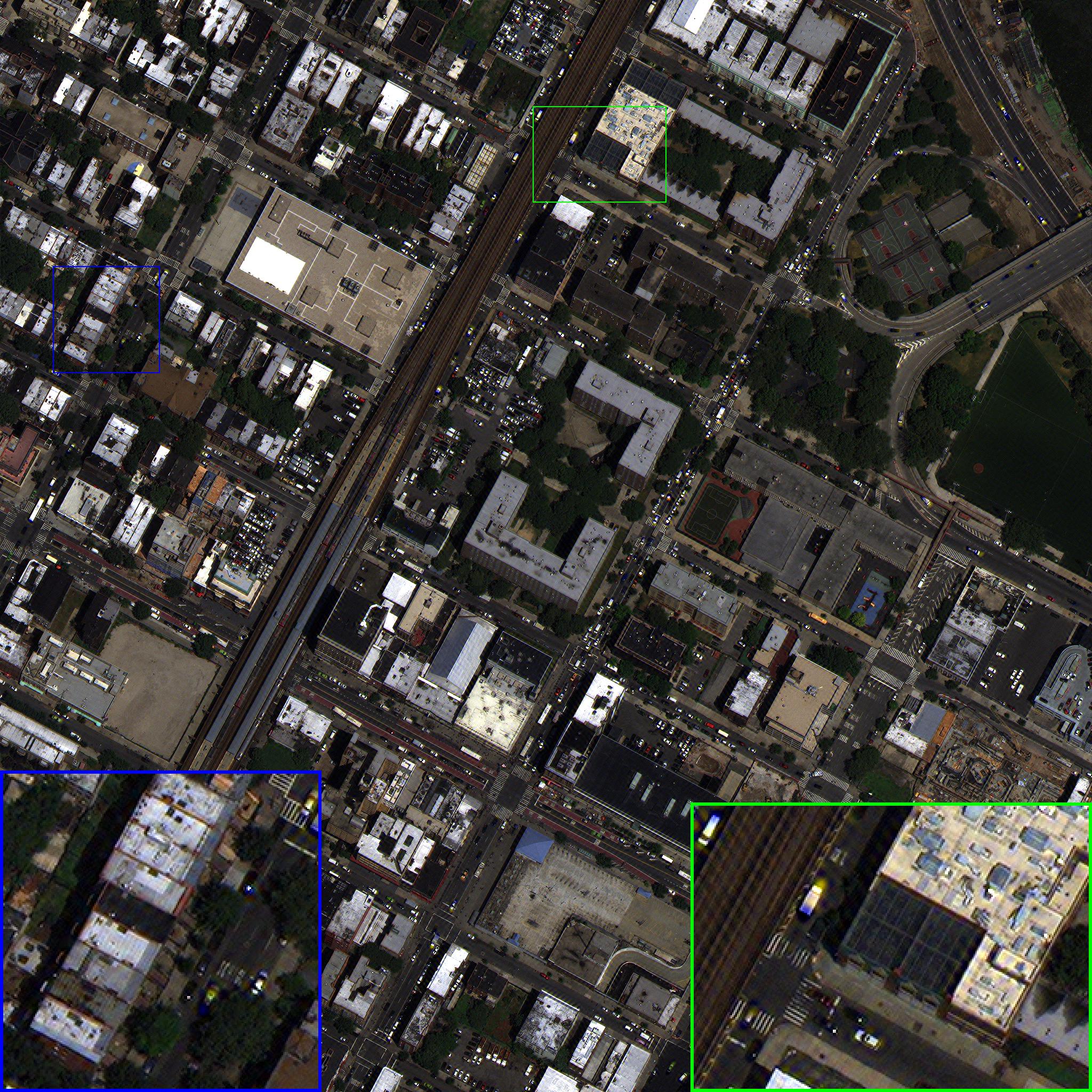}
	\subcaption{{C-GSA}}
\end{subfigure}
\hfill
\begin{subfigure}[t]{.24\textwidth}
	\centering
\includegraphics[width=\textwidth]{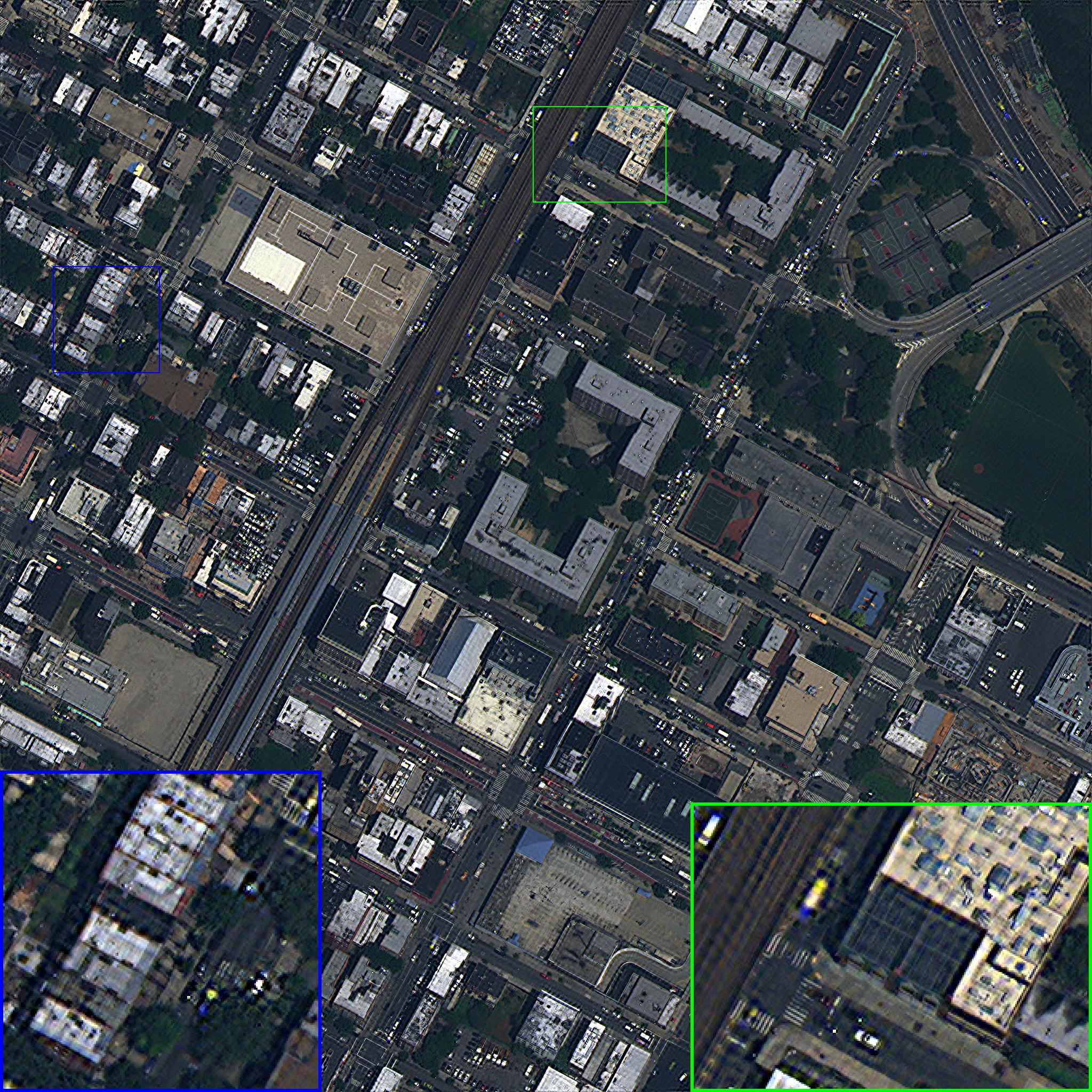}
	\subcaption{{SR-D}}
\end{subfigure}
\hfill
\begin{subfigure}[t]{.24\textwidth}
	\centering
\includegraphics[width=\textwidth]{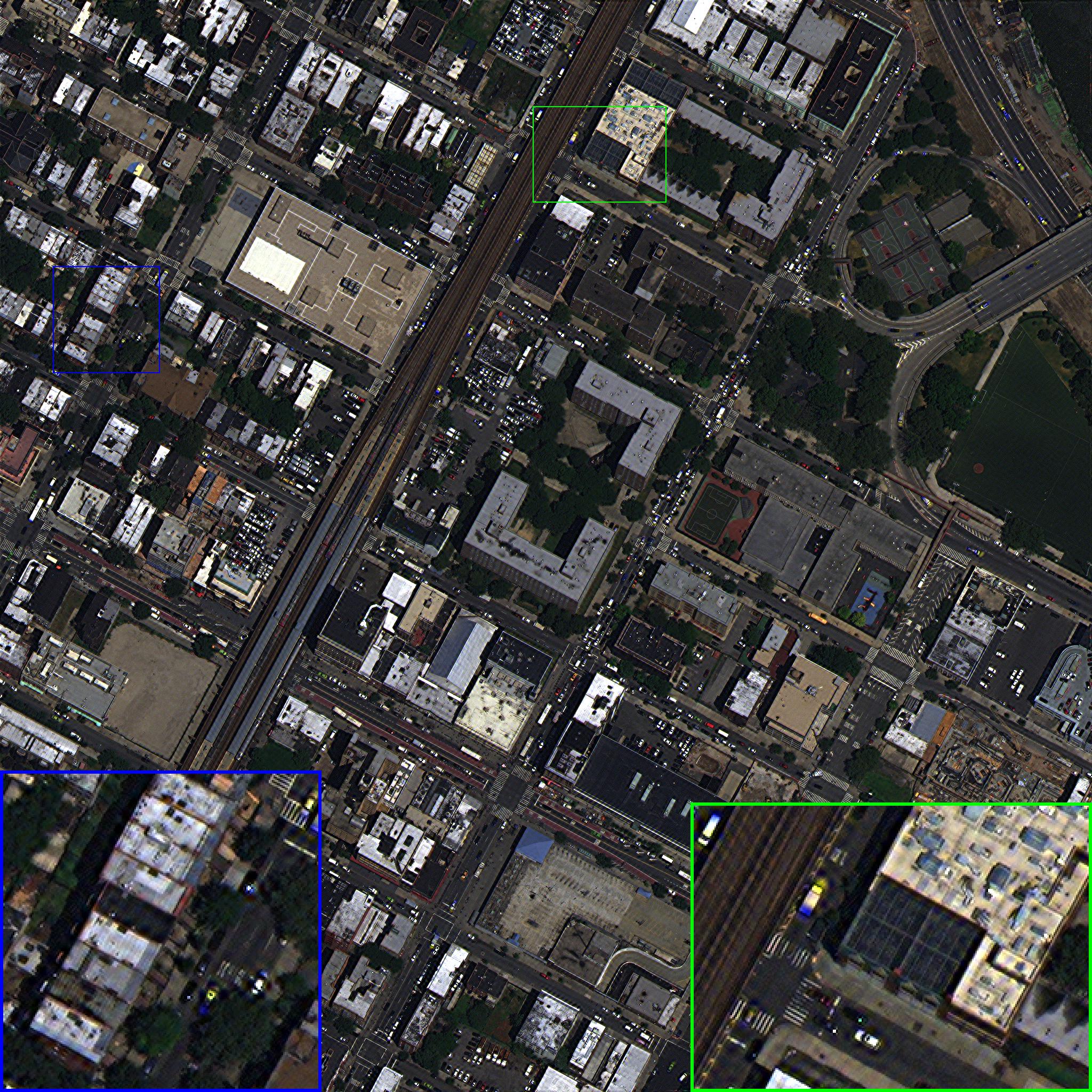}
	\subcaption{{MTF-GLP-HPM-R}}
\end{subfigure}
\hfill
\begin{subfigure}[t]{.24\textwidth}
	\centering
\includegraphics[width=\textwidth]{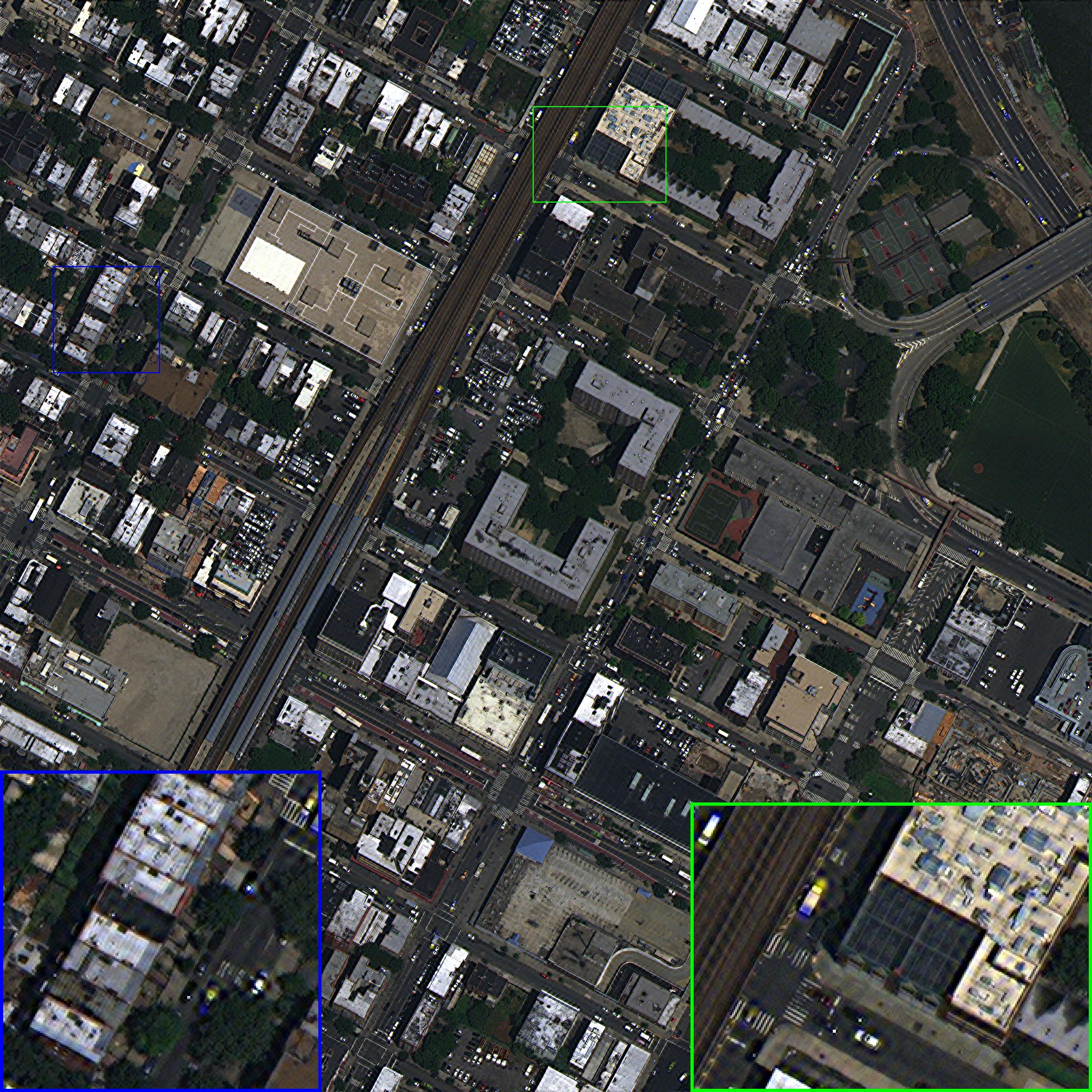}
	\subcaption{{MTF-GLP-FS}}
\end{subfigure}
\hfill
\begin{subfigure}[t]{.24\textwidth}
	\centering
\includegraphics[width=\textwidth]{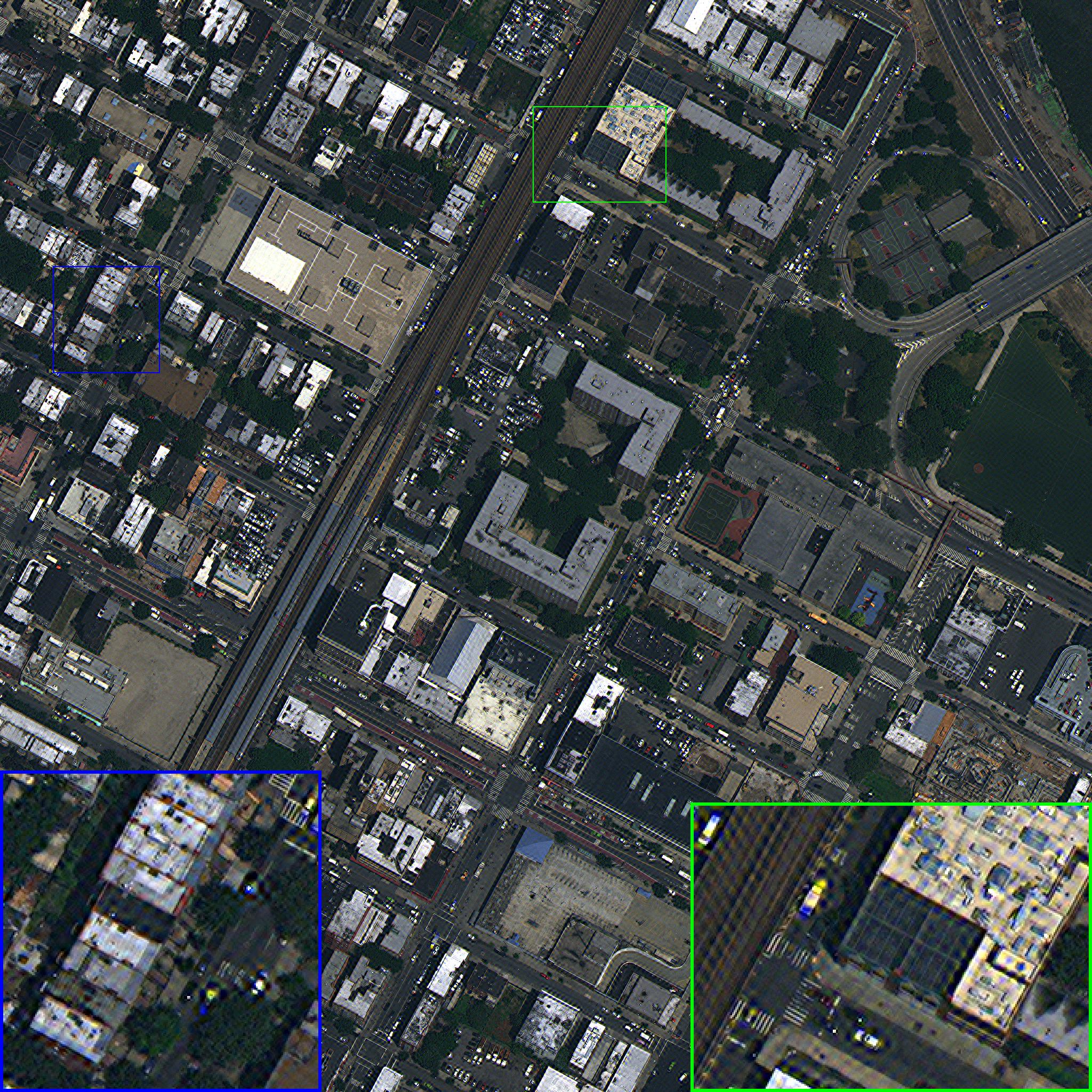}
	\subcaption{{TV}}
\end{subfigure}
\hfill
\begin{subfigure}[t]{.24\textwidth}
	\centering
\includegraphics[width=\textwidth]{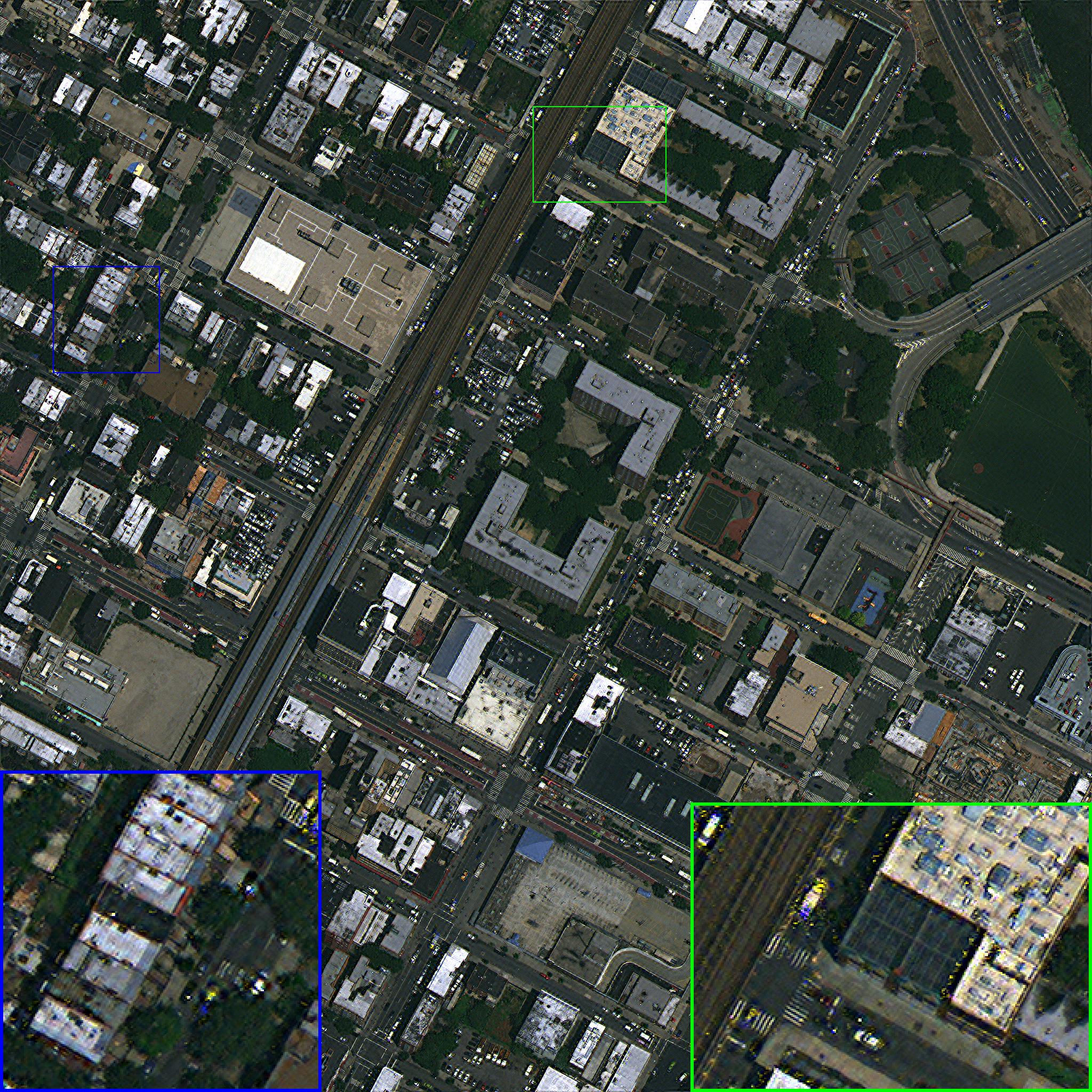}
	\subcaption{{PanNet}}
\end{subfigure}
\hfill
\begin{subfigure}[t]{.24\textwidth}
	\centering
\includegraphics[width=\textwidth]{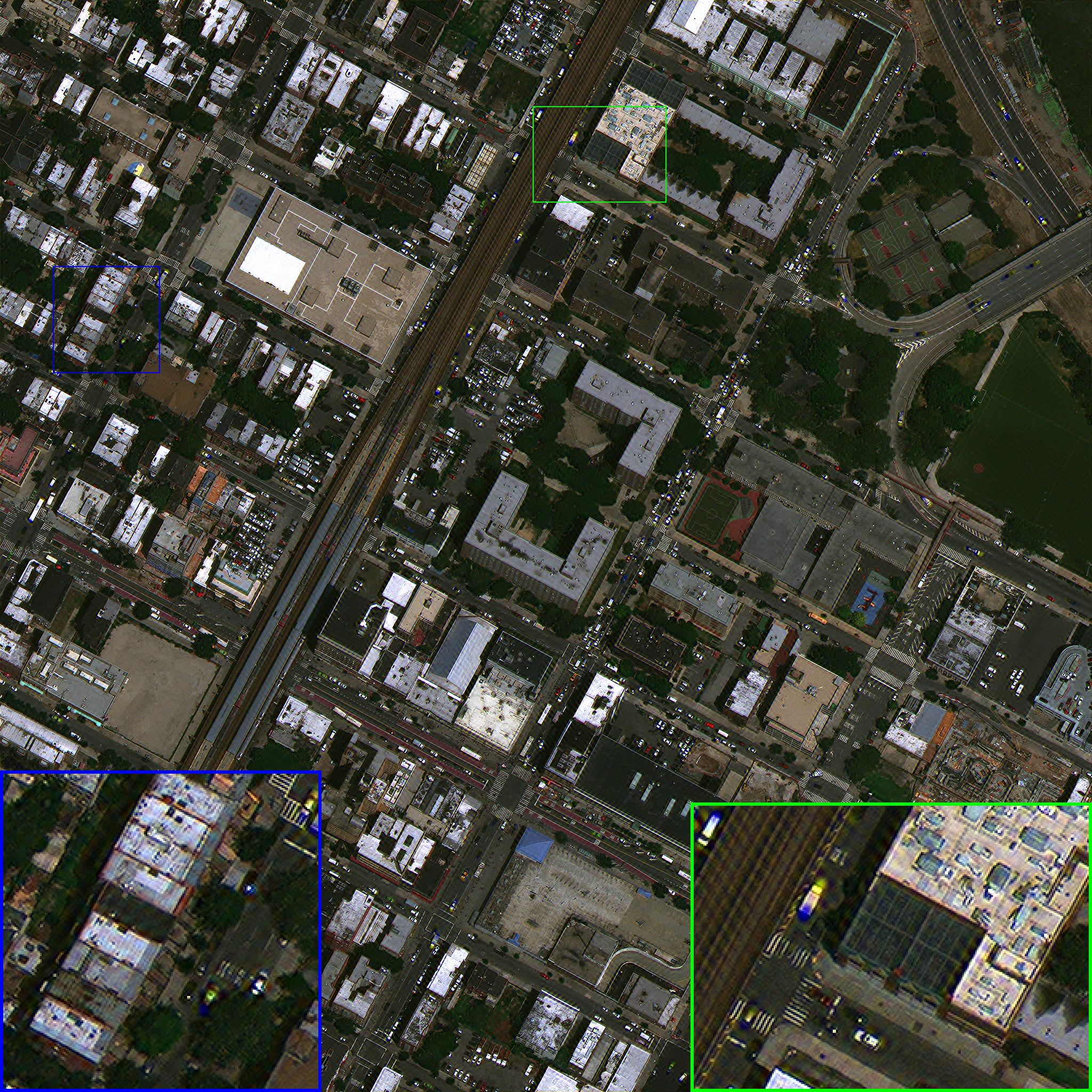}
	\subcaption{{DRPNN}}
\end{subfigure}
\hfill
\begin{subfigure}[t]{.24\textwidth}
	\centering
\includegraphics[width=\textwidth]{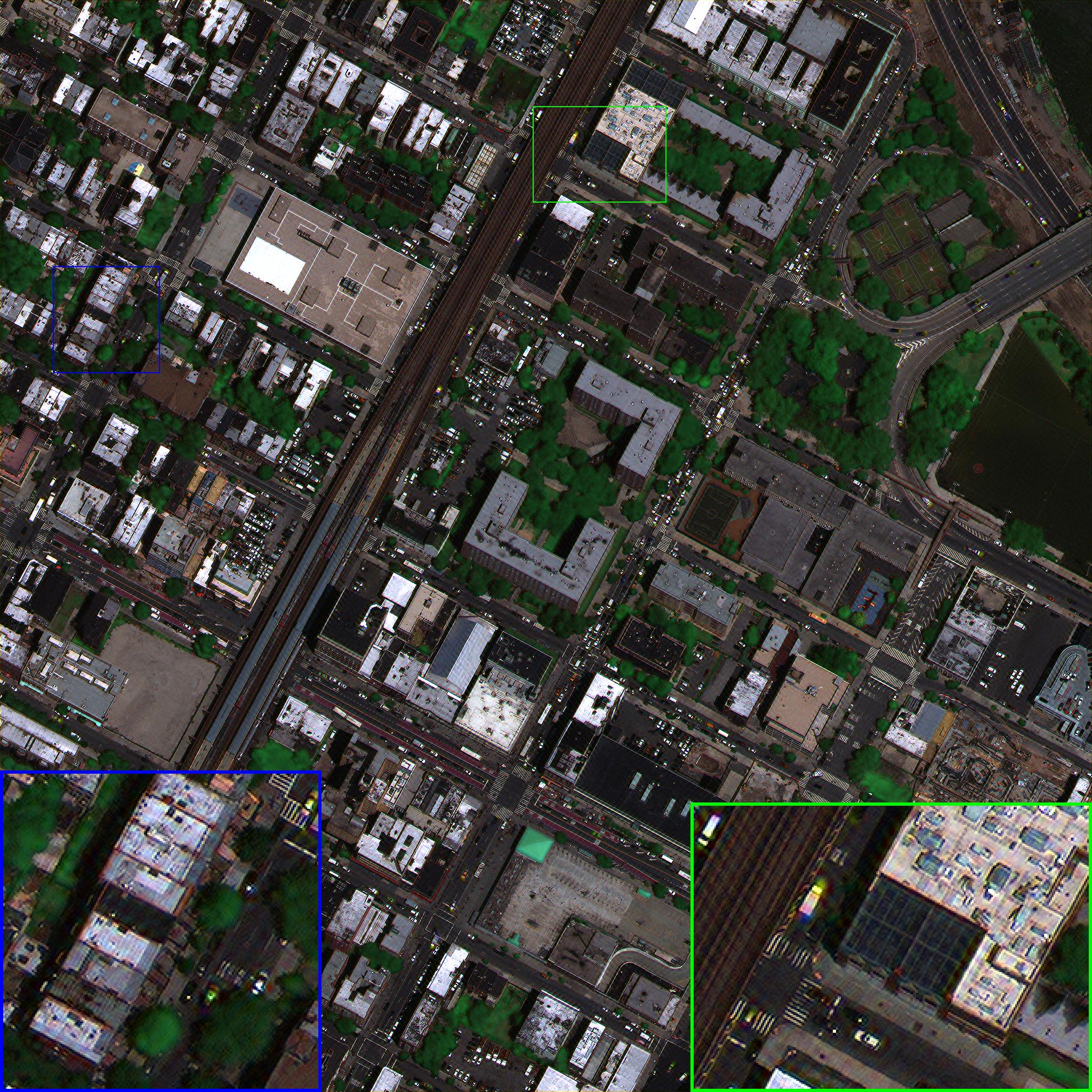}
	\subcaption{{MSDCNN}}
\end{subfigure}
\hfill
\begin{subfigure}[t]{.24\textwidth}
	\centering
\includegraphics[width=\textwidth]{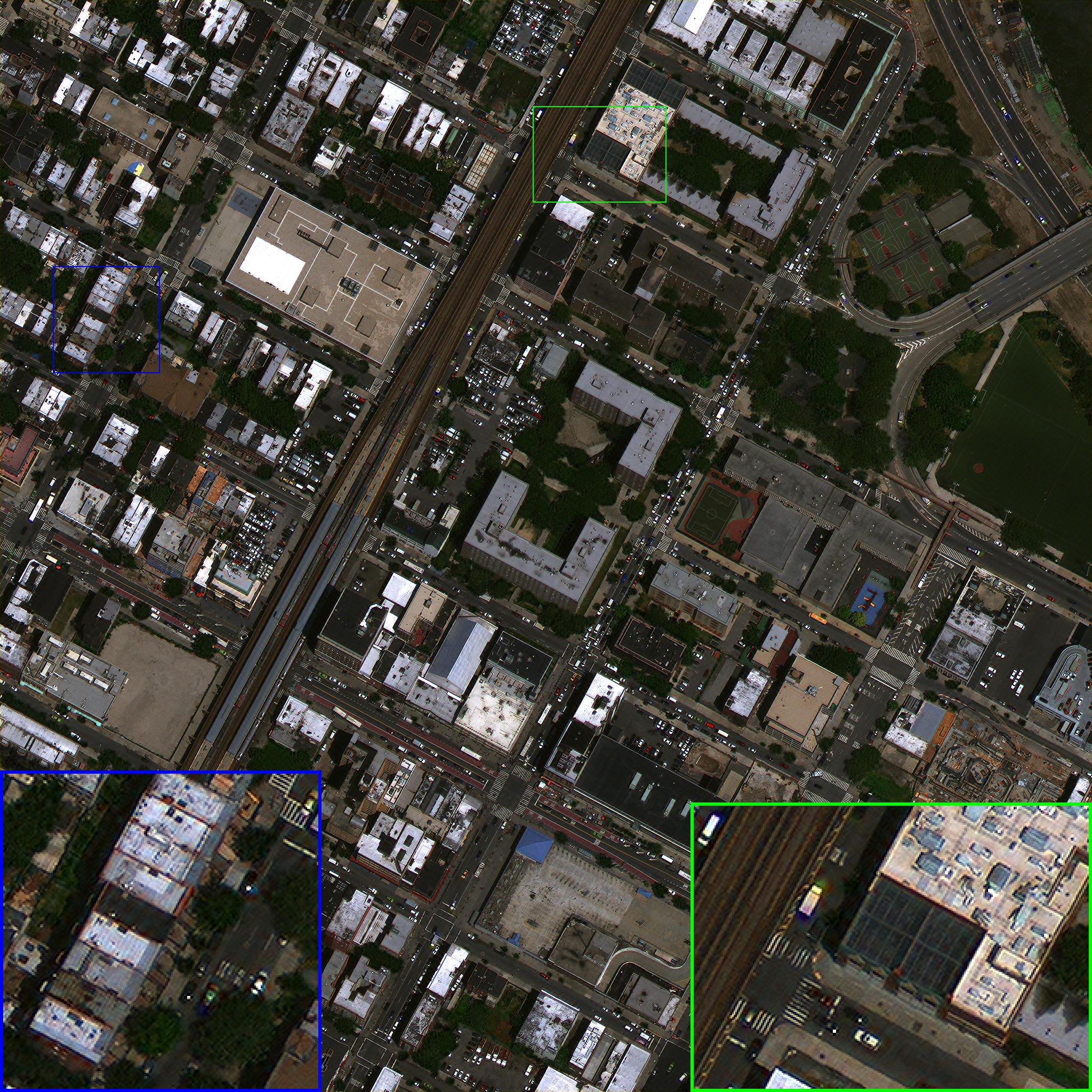}
	\subcaption{{BDPN}}
\end{subfigure}
\hfill
\begin{subfigure}[t]{.24\textwidth}
	\centering
\includegraphics[width=\textwidth]{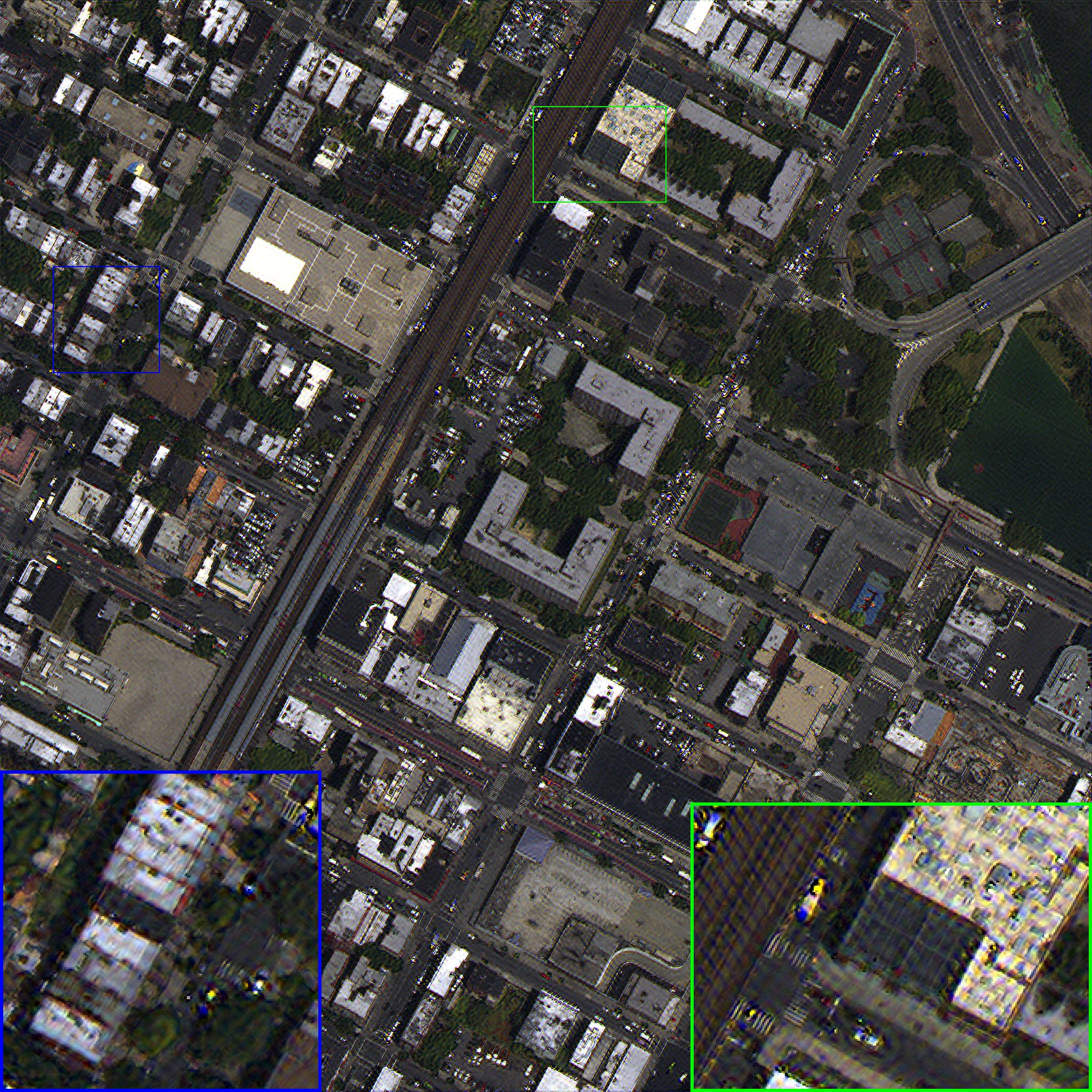}
	\subcaption{{DiCNN}}
\end{subfigure}
\hfill
\begin{subfigure}[t]{.24\textwidth}
	\centering
\includegraphics[width=\textwidth]{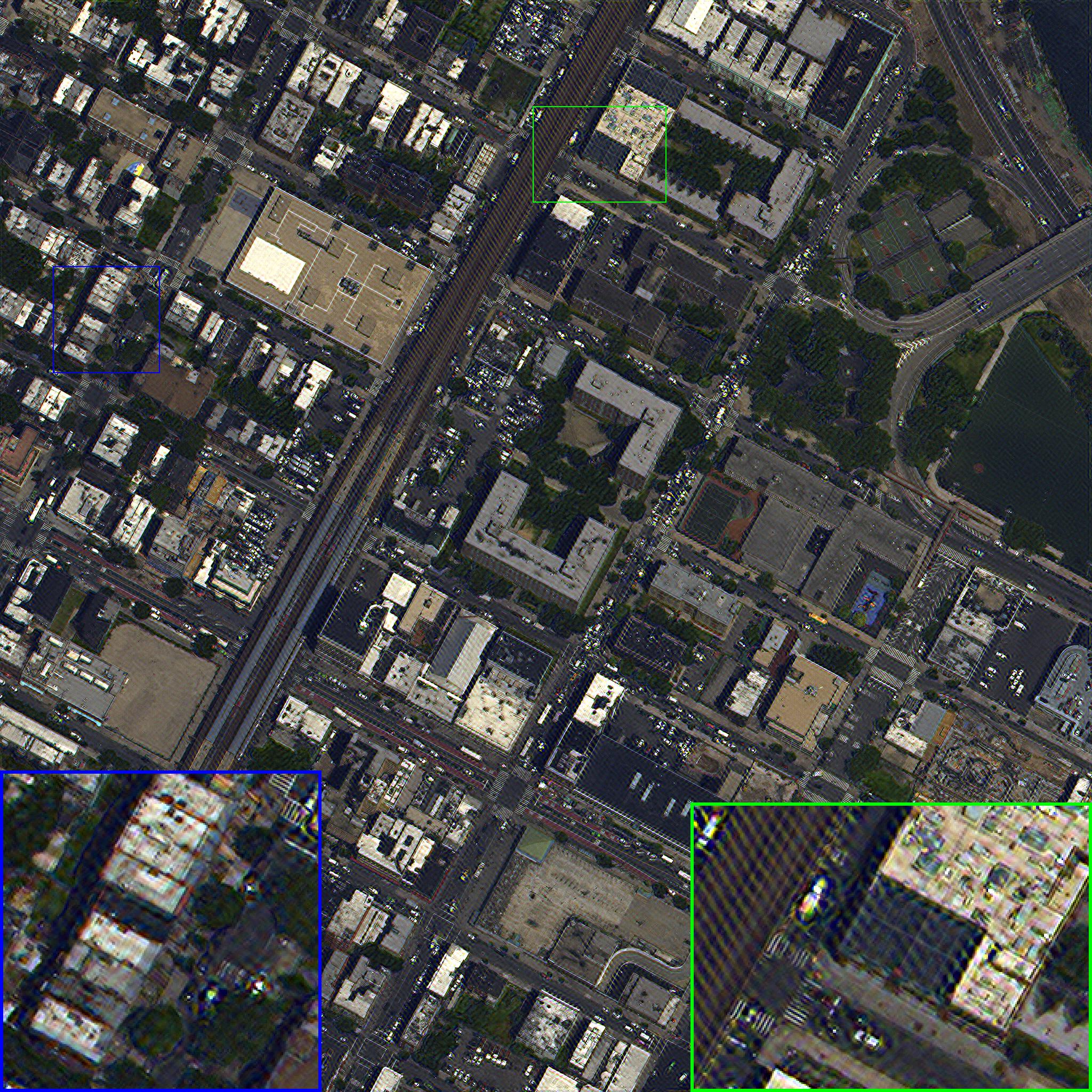}
	\subcaption{{PNN}}
\end{subfigure}
\hfill
\begin{subfigure}[t]{.24\textwidth}
	\centering
\includegraphics[width=\textwidth]{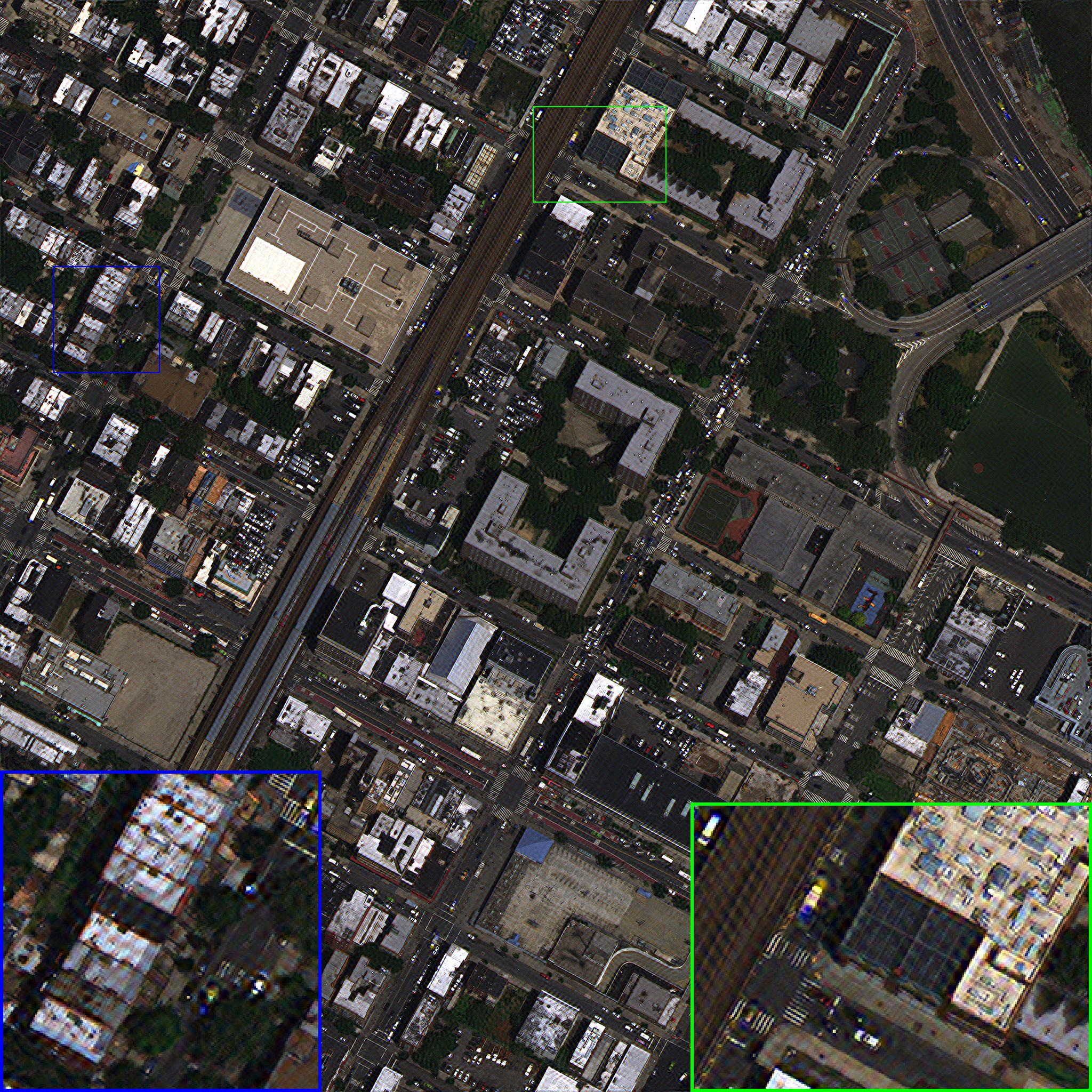}
	\subcaption{{APNN}}
\end{subfigure}
\hfill
\begin{subfigure}[t]{.24\textwidth}
	\centering
\includegraphics[width=\textwidth]{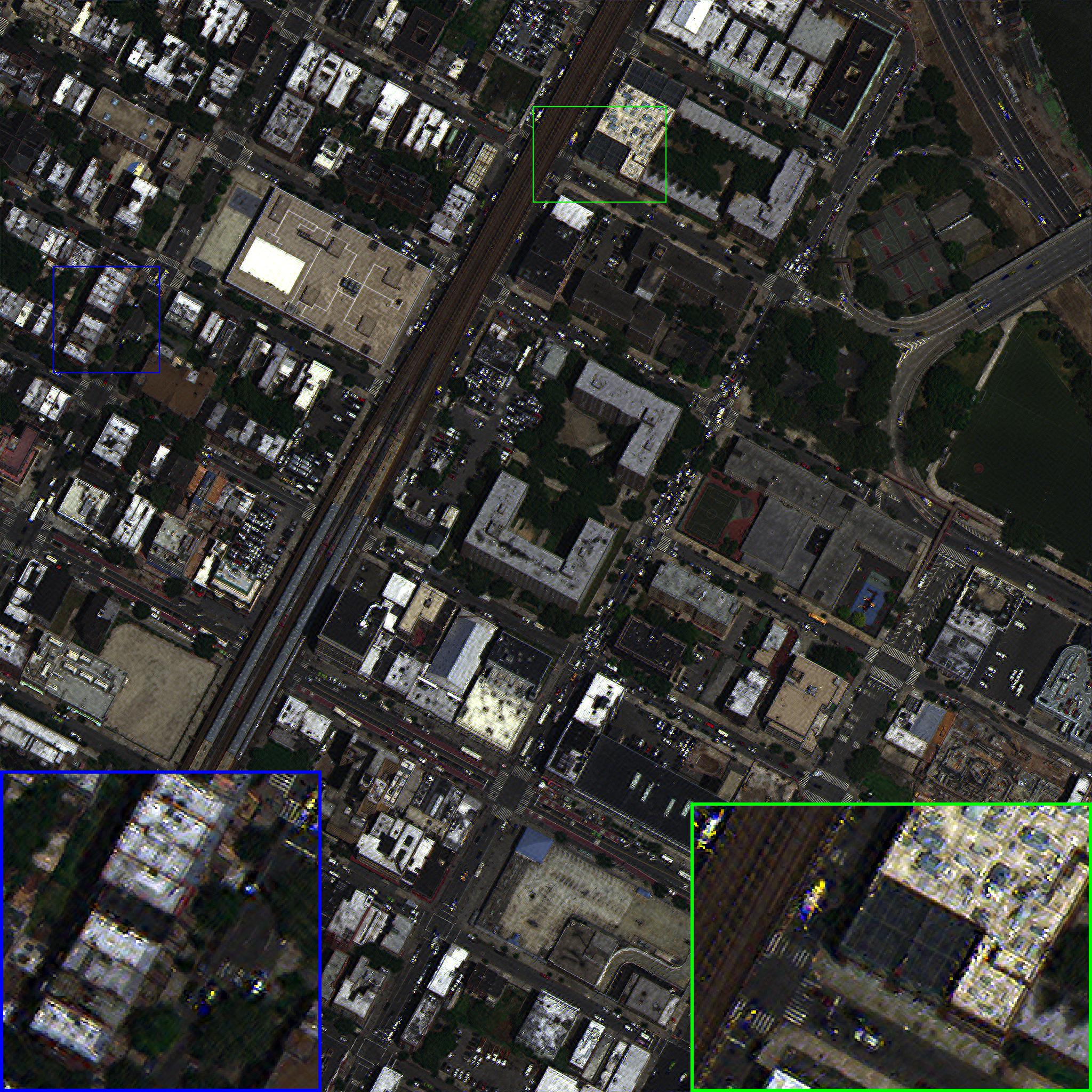}
	\subcaption{{FusionNet}}
\end{subfigure}
\hfill
\begin{subfigure}[t]{.24\textwidth}
	\centering
\includegraphics[width=\textwidth]{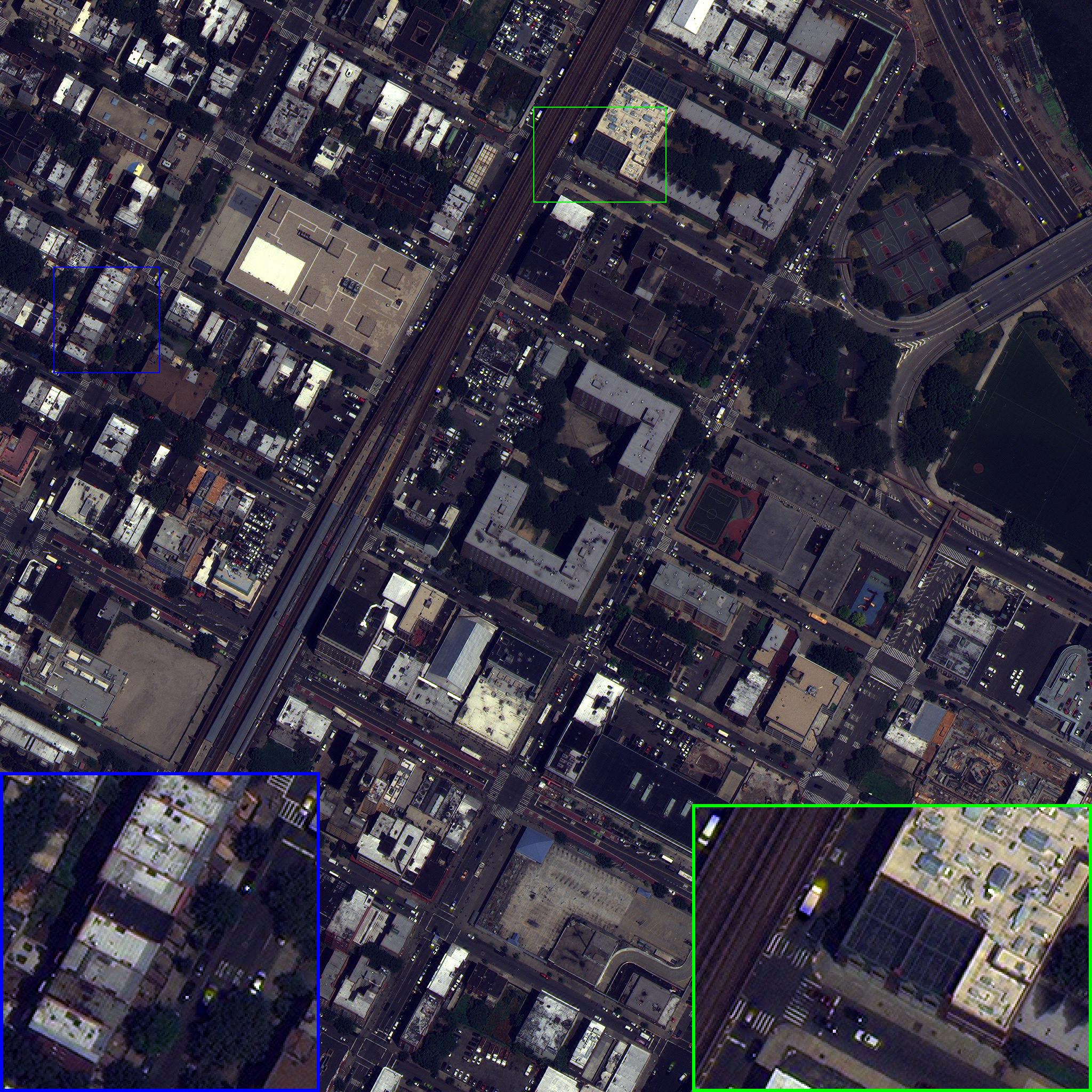}
	\subcaption{{PCS}}
\end{subfigure}
\hfill
\begin{subfigure}[t]{.24\textwidth}
	\centering
\includegraphics[width=\textwidth]{figure/PCS.jpg}
	\subcaption{{PMRA}}
\end{subfigure}
\caption{Comparisons of the different Pan-sharpening methods\cite{deng2022machine}   for WV3 New York dataset.\label{fig:Comparisons of the different pansharpening methods for WV3 New York dataset.}}
\end{figure*}

According to Figure~\ref{fig:Comparisons of the different pansharpening methods for WV3 New York dataset.}, our methods, PCS and PMRA, are clear and the color is realistic, while there are strange textures on the results of PanNet, DRPNN,  MSDCNN, DiCNN, PNN, APNN, FusionNet.

\subsection{Synthetic experiments: performance comparison with Chikusei data}
\subsubsection{Generation of the simulated data}
We implemented two experiments. In the first experiment, we used equal weights for all bands to generate the PAN image and employed 2-scale mean down-sampling to generate the low spatial resolution HS image. Since the resolution of first experiment was high, we also used it for visual demonstration. In the second experiment, we uniformly sampled weights from 0.1 to 1 and linearly normalized the weights to sum to 1, and used these normalized weights to generate the PAN image. We then employed 4-scale mean down-sampling to generate the low spatial resolution HS image. 

We consider the $s=1$  situation. We choose the number 60,40, and 21 bands to represent the red, green, and blue(RGB), spectral channels for rendering real color images. Additionally, we select the number 50,27, and 11 bands to represent orange, azure, and purple(OAP) for rendering synthetic images. 


\begin{figure*}[]
\centering
\captionsetup[subfigure]{font=tiny}
\begin{minipage}{\textwidth}
\hfill
\begin{subfigure}{.19\textwidth}
	\centering
\includegraphics[width=\textwidth]{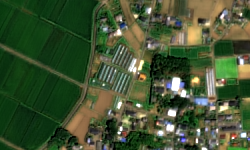}
	\caption{{PCS RGB}}
\end{subfigure}
\hfill
\begin{subfigure}{.19\textwidth}
	\centering
\includegraphics[width=\textwidth]{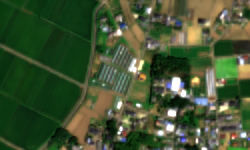}
	\caption{GSA RGB}
\end{subfigure}
\hfill
\begin{subfigure}{.19\textwidth}
	\centering
\includegraphics[width=\textwidth]{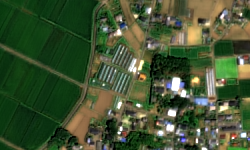}
	\caption{PMRA RGB}
\end{subfigure}
\hfill
\begin{subfigure}{.19\textwidth}
	\centering
\includegraphics[width=\textwidth]{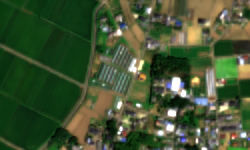}
	\caption{MTF-GLP-CBD RGB}
\end{subfigure}
\hfill
\begin{subfigure}{.19\textwidth}
	\centering
\includegraphics[width=\textwidth]{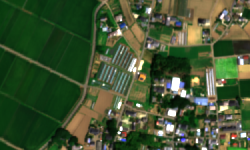}
	\caption{Ground-Truth RGB}
\end{subfigure}
\hfill
\begin{subfigure}{.19\textwidth}
	\centering
\includegraphics[width=\textwidth]{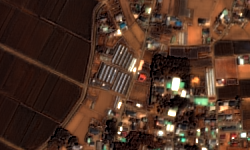}
	\caption{{PCS OAP}}
\end{subfigure}
\hfill
\begin{subfigure}{.19\textwidth}
	\centering
\includegraphics[width=\textwidth]{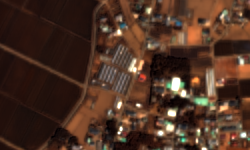}
	\caption{{GSA OAP}}
\end{subfigure}
\hfill
\begin{subfigure}{.19\textwidth}
	\centering
\includegraphics[width=\textwidth]{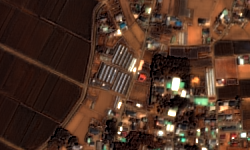}
	\caption{{PMRA OAP}}
\end{subfigure}
\hfill
\begin{subfigure}{.19\textwidth}
	\centering
\includegraphics[width=\textwidth]{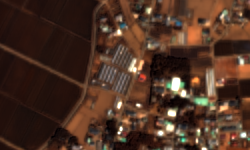}
	\caption{{MTF-GLP-CBD OAP}}
\end{subfigure}
\hfill
\begin{subfigure}{.19\textwidth}
	\centering
\includegraphics[width=\textwidth]{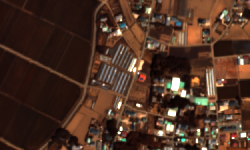}
	\caption{{Ground-Truth OAP}}
\end{subfigure}
\caption{Comparisons of the different synthetic methods with ground truth.\label{fig:compare synthetic and GT} RGB represents the red green blue channel and OAP represents the other selected three channels.}
\end{minipage}
\end{figure*}


 \begin{table}[ht!]
 \caption{Chikusei synthetic experiments}
  \centering
    \resizebox{0.48\textwidth}{!}{
  \begin{subtable}{\textwidth}
    \centering
    \subcaption{Equally weighted PAN and 2-scale mean down-sampling HS\label{tab:Chikusei synthetic experiment}}
    \begin{tabular}{ccccccc}
\hline
\multirow{2}{*}{Method}   
    &{Consistent}    &{Spatial}  & Spetral &\multirow{2}{*}{RMSE}& \multirow{2}{*}{$\Ab^{-}\Ab$} & \\ 
       &RMSE($\downarrow$)   &RMSE($\downarrow$) & RMSE($\downarrow$)& ($\downarrow$) & &     \\ \hline
MTF-GLP-CBD   & 0.00 & 0.00 & 47.12    &   79.25   & 1.00       \\ \hline
GSA           & 0.00 & 0.00 & 47.12    &   79.25   & 1.00      \\ \hline
PCS           & 0.00 & 0.00 & 55.70    &   93.93   & 1.00      \\ \hline
PMRA          & 0.00 & 0.00 & 55.70    &   93.93   & 1.00      \\ \hline
\end{tabular}
\end{subtable}}%
  \hfill
  \resizebox{0.48\textwidth}{!}{
  \begin{subtable}{\textwidth}
    \centering
        \subcaption{Randomly weighted PAN and 4-scale mean down-sampling HS\label{tab:Chikusei synthetic experiment random}}
    \begin{tabular}{ccccccc}
\hline
\multirow{2}{*}{Method}   
    &{Consistent}    &{Spatial}  & Spetral &\multirow{2}{*}{RMSE}& \multirow{2}{*}{$\Ab^{-}\Ab$} & \\ 
       &RMSE($\downarrow$)   &RMSE($\downarrow$) & RMSE($\downarrow$)& ($\downarrow$) & &     \\ \hline
MTF-GLP-CBD   & 132.46 & 89.48 & 41.67    &   186.02   & 0.92       \\ \hline
GSA           & 132.46 & 0.00 & 166.33    &   154.85   & 1.00      \\ \hline
PCS           & 0.00 & 0.00 & 18.14    &   186.13   & 1.00      \\ \hline
PMRA          & 0.00 & 0.00 & 18.14    &   186.13   & 1.00      \\ \hline
\end{tabular}
\end{subtable}}
\end{table}

\subsubsection{Existence of the solution}

According to table~\ref{tab:Chikusei synthetic experiment}, the $consistent\ RMSE = 0$ implies the solutions could exist. 
According to table~\ref{tab:Chikusei synthetic experiment random}, the $consistent\ RMSE$ for GSA and MTF-GLP-CBD are higher than that for PCS and PMRA since PCS and PMRA employ down-sampling enhancement techniques.
\subsubsection{Recover of the solution}

The recovered solutions for the first experiment are shown in Figure~\ref{fig:compare synthetic and GT}. According to Table~\ref{tab:Chikusei synthetic experiment} and Table~\ref{tab:Chikusei synthetic experiment random}, all the methods in this experiment exhibit  general invertibility for $\Ab$ and $\Ab^{-}$ except for MTF-GLP-CBD in the second experiment. Additionally, the spatial RMSE values for all four methods are zero except for the MTF-GLP-CBD in the second experiment. However, the spectral RMSE are non-zero, indicating that the down-sampling and up-sampling matrices are not matched. This mismatch arises because we employ bilinear interpolation for down-sampling and up-sampling operators to prioritize visual performance. The similar results of GSA and MTF-GLP-CBD in Table~\ref{tab:Chikusei synthetic experiment} may be due to the experiment setting (the construction of the PAN image).

Regarding RMSE, the proposed methods exhibit higher values compared to the GSA and MTF-GLP-CBD methods. This phenomenon may due to a better meet of  $\Ab^{-}$ and RMSE extracted by the GSA and MTF-GLP-CBD in these synthetic experiments. Since the prior for $\Ab^{-}$ is designed to optimize visual quality, a minor deficiency in the proposed methods' RMSE is deemed acceptable.

The image results show that methods derived from the generalized inverse have good performance, while the proposed PCS and PMRA exhibit better visual sharpness.

\subsection{Ablation study: performance comparison with PAirMax data}
We used nine scenes in this dataset. The quantitative results reflect  analyses from all nine scenes, while the qualitative results are based on experiments conducted using the GE Tren Urb dataset. We used the MS\_LR in the FF path to conduct the experiment. The mean down-sampling and nearest neighborhood up-sampling (which are matched) were selected for the algorithms and calculations of the spectral RMSE.

\subsubsection{Quantitative results}
DSE is used to denote down-sampling enhancement.

According to table~\ref{tab:GE Lond Urb},\ref{tab:GE Tren Urb},\ref{tab:W2 Miam Mix},\ref{tab:W2 Miam Urb},\ref{tab:W3 Muni Mix},\ref{tab:W3 Muni Nat},\ref{tab:W3 Muni Urb},\ref{tab:W4 Mexi Nat},\ref{tab:W4 Mexi Urb}, the quantitative result shows that GSA and PCS/PMRA methods with down-sampling enhancement are relatively better than MTF-GLP-CBD and vanilla GSA, PCS, and PMRA methods.  The three losses(consistent RMSE, spatial RMSE, spectral RMSE) are all zero for the GSA and PCS/PMRA methods with down-sampling enhancement given $\Ab^{-}\Ab=1$.


 \begin{table}[ht!]
 \caption{Performance comparison on PAirMax data}
  \centering
    \resizebox{0.48\textwidth}{!}{
  \begin{subtable}{\textwidth}
    \centering
    \subcaption{GE Lond Urb\label{tab:GE Lond Urb}}
    \begin{tabular}{ccccccc}
\hline
\multirow{2}{*}{Method}   & \multirow{2}{*}{DSE}
    &Consistent    &Spatial  & Spetral & \multirow{2}{*}{$\Ab^{-}\Ab$} & \\ 
  & 
    &RMSE($\downarrow$)   &RMSE($\downarrow$) & RMSE($\downarrow$)&  &     \\ \hline
MTF-GLP-CBD  & $\times$ & 37.36 & 37.62          & 0.00                   & 0.93     & \\ \hline
GSA &$\times$ & 37.36 & 0.00             & 44.14                   & 1.00     & \\ \hline
PCS& $\times$&  37.36 & 0.00             & 36.13                   & 1.00     & \\ \hline
PMRA &$\times$ &  37.36 &  37.36        & 0.00                   & 1.00     & \\ \hline
MTF-GLP-CBD &$\checkmark$& 0.00 & 5.17            & 0.00                   & 0.93     & \\ \hline
GSA  &$\checkmark$& 0.00 & 0.00             & 0.00                   & 1.00     & \\ \hline
PCS/PMRA&$\checkmark$ & 0.00 &  0.00         & 0.00                   & 1.00     & \\ \hline
\end{tabular}
\end{subtable}}%
  \hfill
  \resizebox{0.48\textwidth}{!}{
  \begin{subtable}{\textwidth}
    \centering
        \subcaption{GE Tren Urb\label{tab:GE Tren Urb}}
    \begin{tabular}{ccccccc}
\hline
\multirow{2}{*}{Method}   & \multirow{2}{*}{DSE}
    &Consistent    &Spatial  & Spetral & \multirow{2}{*}{$\Ab^{-}\Ab$} & \\ 
  & 
    &RMSE($\downarrow$)   &RMSE($\downarrow$) &  RMSE($\downarrow$)&  &     \\ \hline
MTF-GLP-CBD  & $\times$ & 41.85 & 41.96          & 0.00                   & 0.96     & \\ \hline
GSA &$\times$ & 41.85 & 0.00             & 53.00                   & 1.00     & \\ \hline
PCS& $\times$&  41.85 & 0.00             & 53.21                   & 1.00     & \\ \hline
PMRA &$\times$ &  41.85 &  41.85         & 0.00                   & 1.00     & \\ \hline
MTF-GLP-CBD &$\checkmark$& 0.00 & 3.51            & 0.00                   & 0.96     & \\ \hline
GSA  &$\checkmark$& 0.00 & 0.00             & 0.00                   & 1.00     & \\ \hline
PCS/PMRA&$\checkmark$ & 0.00 &  0.00         & 0.00                   & 1.00     & \\ \hline
\end{tabular}
  \end{subtable}}
   \hfill
  \resizebox{0.48\textwidth}{!}{
  \begin{subtable}{\textwidth}
    \centering
        \subcaption{W2 Miam Mix\label{tab:W2 Miam Mix}}
\begin{tabular}{ccccccc}
\hline
\multirow{2}{*}{Method}   & \multirow{2}{*}{DSE}
    &Consistent    &Spatial  & Spetral & \multirow{2}{*}{$\Ab^{-}\Ab$} & \\ 
  & 
    &RMSE($\downarrow$)   &RMSE($\downarrow$) & RMSE($\downarrow$)&  &     \\ \hline
MTF-GLP-CBD  & $\times$ & 26.94 & 27.04          & 0.00                   & 0.96     & \\ \hline
GSA &$\times$ & 26.94 & 0.00             & 27.49                   & 1.00     & \\ \hline
PCS& $\times$&  26.94 & 3.45             & 26.31                   & 1.05     & \\ \hline
PMRA &$\times$ &  26.94 &  27.12        & 0.00                   & 1.05     & \\ \hline
MTF-GLP-CBD &$\checkmark$& 0.00 & 2.49            & 0.00                   & 0.96     & \\ \hline
GSA  &$\checkmark$& 0.00 & 0.00             & 0.00                   & 1.00     & \\ \hline
PCS/PMRA&$\checkmark$ & 0.00 &  3.45         & 0.00                   & 1.05     & \\ \hline
\end{tabular}
\end{subtable}}
   \hfill
  \resizebox{0.48\textwidth}{!}{
  \begin{subtable}{\textwidth}
    \centering
        \subcaption{W2 Miam Urb\label{tab:W2 Miam Urb}}
\begin{tabular}{ccccccc}
\hline
\multirow{2}{*}{Method}   & \multirow{2}{*}{DSE}
    &Consistent    &Spatial  & Spetral & \multirow{2}{*}{$\Ab^{-}\Ab$} & \\ 
  & 
    &RMSE($\downarrow$)   &RMSE($\downarrow$) & RMSE($\downarrow$)&  &     \\ \hline
MTF-GLP-CBD  & $\times$ & 46.77 & 46.85          & 0.00                   & 0.97     & \\ \hline
GSA &$\times$ & 46.77 & 0.00             & 41.25                   & 1.00     & \\ \hline
PCS& $\times$&  46.77 & 0.00             & 48.78                   & 1.00     & \\ \hline
PMRA &$\times$ &  46.77 &  46.77         & 0.00                   & 1.00     & \\ \hline
MTF-GLP-CBD &$\checkmark$& 0.00 & 3.17             & 0.00                   & 0.97     & \\ \hline
GSA  &$\checkmark$& 0.00 & 0.00             & 0.00                   & 1.00     & \\ \hline
PCS/PMRA&$\checkmark$ & 0.00 &  0.00         & 0.00                   & 1.00     & \\ \hline
\end{tabular}
\end{subtable}}
\hfill
  \resizebox{0.48\textwidth}{!}{
  \begin{subtable}{\textwidth}
    \centering
        \subcaption{W3 Muni Mix\label{tab:W3 Muni Mix}}
\begin{tabular}{ccccccc}
\hline
\multirow{2}{*}{Method}   & \multirow{2}{*}{DSE}
    &Consistent    &Spatial  & Spetral & \multirow{2}{*}{$\Ab^{-}\Ab$} & \\ 
  & 
    &RMSE($\downarrow$)   &RMSE($\downarrow$) & RMSE($\downarrow$)&  &     \\ \hline
MTF-GLP-CBD  & $\times$ & 13.58 & 13.64          & 0.00                   & 0.96     & \\ \hline
GSA &$\times$ & 13.58 & 0.00             & 16.43                   & 1.00     & \\ \hline
PCS& $\times$&  13.58 & 3.33             & 13.26                   & 1.09     & \\ \hline
PMRA &$\times$ &  13.58 &  13.93         & 0.00                   & 1.09     & \\ \hline
MTF-GLP-CBD &$\checkmark$& 0.00 & 1.45             & 0.00                   & 0.96     & \\ \hline
GSA  &$\checkmark$& 0.00 & 0.00             & 0.00                   & 1.00     & \\ \hline
PCS/PMRA&$\checkmark$ & 0.00 &  3.33         & 0.00                   & 1.09     & \\ \hline
\end{tabular}
\end{subtable}}
\hfill
  \resizebox{0.48\textwidth}{!}{
  \begin{subtable}{\textwidth}
    \centering
        \subcaption{W3 Muni Nat\label{tab:W3 Muni Nat}}
\begin{tabular}{ccccccc}
\hline
\multirow{2}{*}{Method}   & \multirow{2}{*}{DSE}
    &Consistent    &Spatial  & Spetral & \multirow{2}{*}{$\Ab^{-}\Ab$} & \\ 
  & 
    &RMSE($\downarrow$)   &RMSE($\downarrow$) & RMSE($\downarrow$)&  &     \\ \hline
MTF-GLP-CBD  & $\times$ & 5.44 & 5.45          & 0.00                   & 0.98     & \\ \hline
GSA &$\times$ & 5.44 & 0.00             & 9.36                   & 1.00     & \\ \hline
PCS& $\times$&  5.44 & 0.00             & 5.55                   & 1.00     & \\ \hline
PMRA &$\times$ &  5.44 &  5.44         & 0.00                   & 1.00     & \\ \hline
MTF-GLP-CBD &$\checkmark$& 0.00 & 0.34             & 0.00                   & 0.98     & \\ \hline
GSA  &$\checkmark$& 0.00 & 0.00             & 0.00                   & 1.00     & \\ \hline
PCS/PMRA&$\checkmark$ & 0.00 &  0.00         & 0.00                   & 1.00     & \\ \hline
\end{tabular}
\end{subtable}}
\hfill
  \resizebox{0.48\textwidth}{!}{
  \begin{subtable}{\textwidth}
    \centering
        \subcaption{W3 Muni Urb\label{tab:W3 Muni Urb}}
\begin{tabular}{ccccccc}
\hline
\multirow{2}{*}{Method}   & \multirow{2}{*}{DSE}
    &Consistent    &Spatial  & Spetral & \multirow{2}{*}{$\Ab^{-}\Ab$} & \\ 
  & 
    &RMSE($\downarrow$)   &RMSE($\downarrow$) & RMSE($\downarrow$)&  &     \\ \hline
MTF-GLP-CBD  & $\times$ & 26.40 & 26.59             & 0.00                   & 0.94     & \\ \hline
GSA &$\times$ & 26.40 & 0.00             & 31.99                   & 1.00     & \\ \hline
PCS& $\times$&  26.40 & 0.00             & 26.64                   & 1.00     & \\ \hline
PMRA &$\times$ &  26.40 &  26.40         & 0.00                   & 1.00     & \\ \hline
MTF-GLP-CBD &$\checkmark$& 0.00 & 3.48             & 0.00                   & 0.94     & \\ \hline
GSA  &$\checkmark$& 0.00 & 0.00             & 0.00                   & 1.00     & \\ \hline
PCS/PMRA&$\checkmark$ & 0.00 &  0.00         & 0.00                   & 1.00     & \\ \hline
\end{tabular}
\end{subtable}}
\hfill
  \resizebox{0.48\textwidth}{!}{
  \begin{subtable}{\textwidth}
    \centering
        \subcaption{W4 Mexi Nat\label{tab:W4 Mexi Nat}}
\begin{tabular}{ccccccc}
\hline
\multirow{2}{*}{Method}   & \multirow{2}{*}{DSE}
    &Consistent    &Spatial  & Spetral & \multirow{2}{*}{$\Ab^{-}\Ab$} & \\ 
  & 
    &RMSE($\downarrow$)   &RMSE ($\downarrow$)&RMSE($\downarrow$)&  &     \\ \hline
MTF-GLP-CBD  & $\times$ & 13.58 & 13.58             & 0.00                   & 0.99     & \\ \hline
GSA &$\times$ & 13.58 & 0.00             & 14.40                   & 1.00     & \\ \hline
PCS& $\times$&  13.58 & 0.00             & 12.70                   & 1.00     & \\ \hline
PMRA &$\times$ &  13.58 &  13.58         & 0.00                   & 1.00     & \\ \hline
MTF-GLP-CBD &$\checkmark$& 0.00 & 0.21             & 0.00                   & 0.99     & \\ \hline
GSA  &$\checkmark$& 0.00 & 0.00             & 0.00                   & 1.00     & \\ \hline
PCS/PMRA&$\checkmark$ & 0.00 &  0.00         & 0.00                   & 1.00     & \\ \hline
\end{tabular}
\end{subtable}}
\hfill
  \resizebox{0.48\textwidth}{!}{
  \begin{subtable}{\textwidth}
    \centering
        \subcaption{W4 Mexi Urb\label{tab:W4 Mexi Urb}}
\begin{tabular}{ccccccc}
\hline
\multirow{2}{*}{Method}   & \multirow{2}{*}{DSE}
    &Consistent    &Spatial  & Spetral & \multirow{2}{*}{$\Ab^{-}\Ab$} & \\ 
  & 
    &RMSE($\downarrow$)   &RMSE($\downarrow$) & RMSE($\downarrow$)&  &     \\ \hline
MTF-GLP-CBD  & $\times$ & 40.80 & 40.86             & 0.00                   & 0.97     & \\ \hline
GSA &$\times$ & 40.80 & 0.00             & 45.46                   & 1.00     & \\ \hline
PCS& $\times$& 40.80 & 0.00             & 38.53                   & 1.00     & \\ \hline
PMRA &$\times$ & 40.80 &  40.80         & 0.00                   & 1.00     & \\ \hline
MTF-GLP-CBD &$\checkmark$& 0.00 & 2.62             & 0.00                   & 0.97     & \\ \hline
GSA  &$\checkmark$& 0.00 & 0.00             & 0.00                   & 1.00     & \\ \hline
PCS/PMRA&$\checkmark$ & 0.00 &  0.00         & 0.00                   & 1.00     & \\ \hline
\end{tabular}
\end{subtable}}
\end{table}

\subsubsection{Qualitative results}

\begin{figure*}[]
\centering
\captionsetup[subfigure]{font=tiny}
\begin{subfigure}{.45\textwidth}
	\centering
\includegraphics[width=\textwidth]{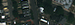}
	\caption{{MS}}
\end{subfigure}
\hfill
\begin{subfigure}{.45\textwidth}
	\centering
\includegraphics[width=\textwidth]{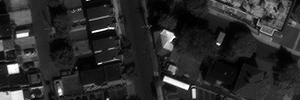}
	\caption{{PAN}}
\end{subfigure}
\hfill
\begin{subfigure}{.45\textwidth}
	\centering
\includegraphics[width=\textwidth]{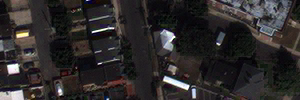}
	\caption{{GSA}}
\end{subfigure}
\hfill
\begin{subfigure}{.45\textwidth}
	\centering
\includegraphics[width=\textwidth]{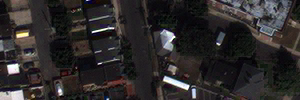}
	\caption{{GSA with DSE}}
\end{subfigure}
\hfill
\begin{subfigure}{.45\textwidth}
	\centering
\includegraphics[width=\textwidth]{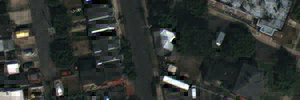}
	\caption{{MTF-GLP-CBD}}
\end{subfigure}
\hfill
\begin{subfigure}{.45\textwidth}
	\centering
\includegraphics[width=\textwidth]{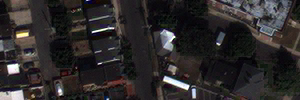}
	\caption{{MTF-GLP-CBD with DSE}}
\end{subfigure}
\hfill
\begin{subfigure}{.45\textwidth}
	\centering
\includegraphics[width=\textwidth]{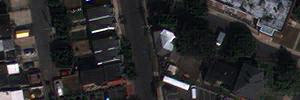}
	\caption{{PCS}}
\end{subfigure}
\hfill
\begin{subfigure}{.45\textwidth}
	\centering
\includegraphics[width=\textwidth]{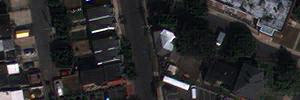}
	\caption{{PCS with DSE}}
\end{subfigure}
\hfill
\begin{subfigure}{.45\textwidth}
	\centering
\includegraphics[width=\textwidth]{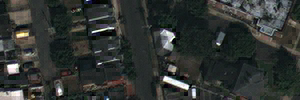}
	\caption{{PMRA}}
\end{subfigure}
\hfill
\begin{subfigure}{.45\textwidth}
	\centering
\includegraphics[width=\textwidth]{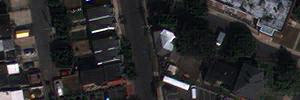}
	\caption{{PMRA with DSE}}
\end{subfigure}

\caption{Comparisons of the different methods for GE Tren Urb dataset.\label{fig:Comparisons of the different  methods for GE_Tren_Urb dataset.}}
\end{figure*}
According to figure~\ref{fig:Comparisons of the different  methods for GE_Tren_Urb dataset.}, we can see that  GSA, GSA with DSE, PCS with DSE, PMRA with DSE, and MTF-GLP-CBD with DSE exhibit relatively better qualitative results. MTF-GLP-CBD and vanilla PMRA appear blurred and sawtooth-blocked. These results are consistent with the quantitative results presented in the tables. Additionally, it shows that down-sampling enhancement is not only beneficial for quantitative results but also effective for improving image appearance.

\subsection{Diffusion-related experiments: performance comparison with PAirMax-RGB data}
There are nine scenes in this dataset. The quantitative results reflect statistical analyses from all nine scenes, while the qualitative results are based on experiments conducted using the W4 Mexi Nat-RGB dataset. 
\subsubsection{Quantitative results}
According to Table~\ref{tab:performance}, The PCS+diffusion method achieves the best performance among all the evaluation measures. The GSA+diffusion method achieves the second-best performance. It demonstrates that employing a diffusion prior can significantly improve the Pan-sharpening results. Since we used average pooling and nearest up-sampling, the down-sampling and up-sampling matrix are matched and therefore spectral RMSE is zero. The similar results of GSA and MTF-GLP-CBD in this case may be due to the experiment setting (the construction of the PAN image).
\begin{table}[htbp]
\centering
\caption{Comparison of different methods on PAirMax-RGB(mean $\pm$ standard deviation). We left appropriate decimal places to facilitate the comparison of these methods.)}
\label{tab:performance}
\resizebox{0.48\textwidth}{!}{
\begin{tabular}{lccccccc}
\hline
Method & RMSE (↓) &  Spatial RMSE (↓) & Spectral RMSE(↓) & PSNR (↑) & SSIM (↑) & SAM (↓) & ERGAS (↓) \\
\hline
PCS           & 0.0046 ± 0.0023 & $0.00 \pm 0.00 $ & $0.00 \pm 0.00$ & 47.58 ± 3.89 & 0.9981 ± 0.0012 & 1.15 ± 0.37 & 9.61 ± 2.78 \\\hline
PCS+diffusion  & \textbf{0.0030} ± 0.0015 & $0.00 \pm 0.00 $ & $0.00 \pm 0.00$ & \textbf{51.26} ± 3.43 & \textbf{0.9990} ± 0.0007 & \textbf{0.81} ± 0.13 & \textbf{6.32} ± 2.05 \\\hline
GSA           & 0.0056 ± 0.0031 & $0.00 \pm 0.00$ & $0.00 \pm 0.00$ & 46.18 ± 4.20 & 0.9974 ± 0.0023 & 1.44 ± 0.42 & 11.28 ± 3.26 \\\hline
GSA+diffusion          & \underline{0.0040} ± 0.0021 & $0.00 \pm 0.00$ & $0.00 \pm 0.00$ & \underline{49.01} ± 3.97 & \underline{0.9984} ± 0.0012 & \underline{1.12} ± 0.30 & \underline{8.14} ± 2.27 \\\hline
MTF-GLP-CBD           & 0.0056 ± 0.0031 & $0.00 \pm 0.00$ & $0.00 \pm 0.00$ & 46.18 ± 4.20 & 0.9974 ± 0.0023 & 1.44 ± 0.42 & 11.28 ± 3.26 \\
\hline
\end{tabular}}
\end{table}

\subsubsection{Qualitative results}
\begin{figure*}[htb]
\centering
\captionsetup[subfigure]{font=tiny}
\begin{subfigure}[t]{.16\textwidth}
	\centering
\includegraphics[width=\textwidth]{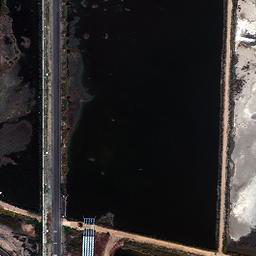}
	\caption{{GT}}
\end{subfigure}
\hfill
\begin{subfigure}[t]{.16\textwidth}
	\centering
\includegraphics[width=\textwidth]{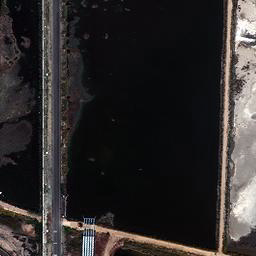}
	\caption{{PCS}}
\end{subfigure}
\hfill
\begin{subfigure}[t]{.16\textwidth}
	\centering
\includegraphics[width=\textwidth]{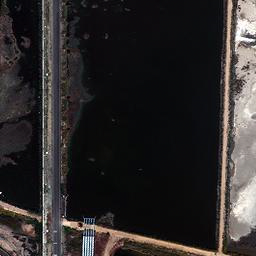}
	\caption{{PCS+diffusion}}
\end{subfigure}
\hfill
\begin{subfigure}[t]{.16\textwidth}
	\centering
\includegraphics[width=\textwidth]{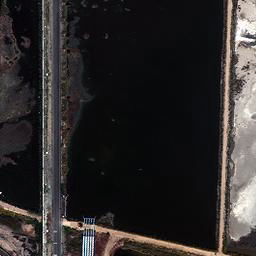}
	\caption{{GSA}}
\end{subfigure}
\hfill
\begin{subfigure}[t]{.16\textwidth}
	\centering
\includegraphics[width=\textwidth]{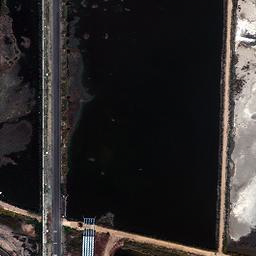}
	\caption{{GSA+diffusion}}
\end{subfigure}
\hfill
\begin{subfigure}[t]{.16\textwidth}
	\centering
\includegraphics[width=\textwidth]{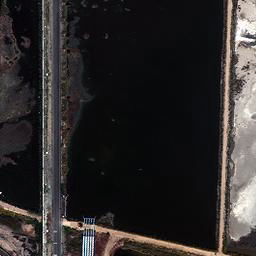}
	\caption{{MTF-GLP-CBD}}
\end{subfigure}
\caption{Comparisons of the different methods for W4 Mexi Nat-RGB dataset.\label{fig:Comparisons of the different  methods for W4_Mexi_Nat-RGB.}}
\end{figure*}

According to Figure~\ref{fig:Comparisons of the different  methods for W4_Mexi_Nat-RGB.}, all the methods demonstrate similar visual performances.
\section{Limitations and discussions}
Although our method demonstrates good results, our consideration of the generalized inverse in the Pan-sharpening problem limits $s=1$ in the prior setting section. We will explore the prior when $s>1$ in future work. Additionally, the spatial response functions and their inverse functions, including down-sampling and up-sampling matrices, utilize bilinear, mean or nearest neighbor functions for approximation. The spatial response functions may need to be accurately estimated, and their generalized inverse may need to be determined precisely. A current limitation is that the current spatial down-sampling matrix is assumed to be fully known. However, the specified spatial down-sampling process and the actual spatial down-sampling process may differ, which would further affect the calculation of the spectral down-sampling matrix to some extent. Since we have deduced the general solution space, in future work, we could let the model learn a proper spatial down-sampling matrix by searching within the prior knowledge provided by a diffusion model. Once this spatial down-sampling matrix is learned, we could derive a more accurate spectral down-sampling matrix. As for the model prior, the relationship to the range of $\Ab^{-}$ requires more dedicated exploration, and the spectral information of object information can be utilized by deep learning models.

Although our methods incorporate diffusion priors,  the diffusion model currently used  is based on ImageNet\cite{deng2009imagenet}, which has only three spectral channels. This limits the applicability of our method. We believe that foundation diffusion models for remote sensing images—which do not yet exist—would be an effective way to extend the utility of our approach.

Although our mathematical models for CS and MRA are elegant, we have not yet considered the impact of noise on the model solution space. Since our method aligns with some existing CS and MRA frameworks, studying the effects of noise could naturally extend these methods into more robust forms. Further research into the influence of noise on our models remains an important area for future investigation.

\section{Conclusion}

In this paper, we utilize generalized inverse theory to understand and solve the matrix representation of the Pan-sharpening problem. By considering the generalized inverse of the matrices, we deduce a representation for the spectral response matrix with the help of the Moore–Penrose inverse. Specifically, we introduce down-sampling enhancement with Moore-Penrose inverse reprojection on the down-sampling matrix.

According to the \cite{radhakrishna1967calculus}, we find a high-spatial and high-spectral image solution space corresponding to the commonly used MRA Pan-sharpening regime using the generalized inverse. Similarly, we derive another high-spatial and high-spectral image solution space corresponding to the commonly used CS Pan-sharpening regime with the help of via the generalized inverse. Specifically, we prove that the GSA method is a special case of our generalized-inverse-based CS framework.

Based on the deduction of the CS and MRA, we propose a prior on the generalized inverse of the spectral response matrix. This yields  algorithms for PCS and PMRA. 

Since people may not use the exact generalized inverse to substitute in the PCS and PMRA regime, we analyze their error. Specifically, we deduce that under down-sample enhancement, PCS and PMRA are equivalent. Besides, we discuss the generalized inverse representation may differ from the oracle solution. The difference depends on the oracle solution, the generalized inverse of the down-sampling matrix, and the generalized inverse of the spectral response matrix.

To further incorporate the prior knowledge of the diffusion model, we substitute the general solution spaces of the generalized inverse form into the evidence lower bound of the diffusion model and optimize the auxiliary matrix $\Wb$ by Score Distillation Sampling to obtain a refined Pan-sharpening solution.

According to the matrix representation of the Pan-sharpening problem, we use consistent RMSE, Spatial RMSE, Spectral RMSE, Inverse Ability, and RMSE  to evaluate the models derived from the generalized inverse theories. In extensive comparison experiments, PCS and PMRA have lowest spatial and spectral RMSE, and their results are clear and sharp. In the synthetic experiments, PCS and PMRA's results are more clear and possess sharp boundaries.  In the real data experiment for ablation study, we show that GSA, PCS, and PMRA with down-sample enhancement are quantitatively better, besides results of MTF-GLP-CBD and PMRA with down-sample enhancement are clear and possess sharp boundaries. The effect of the proposed down-sample enhancement is also validated in both the qualitative and quantitative experiment results. The experiment conducted to validate the effect of the diffusion prior demonstrates that the diffusion prior can significantly improve the performance of PCS and GSA across almost all evaluation measures even though the generated images are visually similar. It demonstrates that after identifying the general solution spaces for CS and MRA methods via generalized inverse theory, we can search within these solution spaces to refine Pan-sharpening techniques.

\section{Data availability statements}
 The hyperspectral data that support the findings of this study are openly available in https://www.sal.t.u-tokyo.ac.jp/hyperdata/. The PAirMax dataset data that support the findings of this study are openly available in 
 
 \noindent https://resources.maxar.com/product-samples/pansharpening-benchmark-dataset.
The WV3 New York dataset can be found in https://github.com/liangjiandeng/DLPan-Toolbox\cite{deng2022machine}.

The authors confirm that the other data supporting the findings of this study are available within the article.

\section*{Acknowledgements}
We thank Xinlin Xie, Cong Liu, and Xuchen Zhang for the discussions on GSA methods. We also thank Fan Wang for sharing practical experience with regularized Bayesian GSA. We thank Xinyang Han for the discussion regarding the projection model.

\bibliography{mybibfile}

\onecolumn
\section*{Appendix}
\subsection{Proofs}
\noindent \textbf{Theorem}~\ref{thm:BY=ZA} We can use linear equations to rewrite the  $\Bb\Yb-\Zb\Ab$ in Kronecker form
, i,e,
\begin{equation}
\begin{bmatrix}
\Zb_{hw\times S}\otimes\Ib_{s\times s},
-\Ib_{hw\times hw}\otimes\Yb^T_{s\times HW}
\end{bmatrix}
\begin{bmatrix}
vec(\Ab)\\
vec(\Bb)
\end{bmatrix}_{Ss+hwHW}
=
0,
\end{equation}
where the Kronecker product $\Zb\otimes\Ib =\begin{bmatrix}
                     z_{1,1}\Ib & z_{1,2}\Ib & \cdots & z_{1,S}\Ib \\
                     z_{2,1}\Ib &  z_{2,2}\Ib & \cdots & z_{2,S}\Ib\\
                     \vdots & \vdots &\ddots& \vdots \\
                     z_{hw,1}\Ib &  z_{hw,2}\Ib & \cdots & z_{hw,S}\Ib
                   \end{bmatrix} $ and $vec(\Ab)= [a_{11},\cdots,a_{S1},a_{12},\cdots,a_{S2},\cdots,a_{1s}\cdots,a_{Ss}]^T.$

 $\Bb\Yb=\Zb\Ab$ has a nonzero solution for $\Bb$ and $\Ab$\\ $\Leftrightarrow$ \begin{itemize}
                                                                                  \item ($HW\ge s$) or ($HW<s$ and $hw<\frac{Ss}{s-HW}$)
                                                                                  \item ($HW<s$ and $hw\ge\frac{Ss}{s-HW}$) and $rank(\begin{bmatrix}
\Zb_{hw\times S}\otimes\Ib_{s\times s},
-\Ib_{hw\times hw}\otimes\Yb^T_{s\times HW}
\end{bmatrix})<min(hws,Ss+hwHW)$.
                                                                                \end{itemize}
\begin{proof} Let $\Mb =\begin{bmatrix}
\Zb_{hw\times S}\otimes\Ib_{s\times s},
-\Ib_{hw\times hw}\otimes\Yb^T_{s\times HW}
\end{bmatrix} $ and $\xb=\begin{bmatrix}
vec(\Ab)\\
vec(\Bb)
\end{bmatrix}_{Ss+hwHW}.$
It yields 
\\$\Mb\xb = 0$ has non-zero solution \\ $\Leftrightarrow$
\begin{itemize}
  \item $hws<Ss+hwHW$
  \item ($hws\ge Ss+hwHW$ and $rank(\Mb)<min(hws,Ss+hwHW)$
\end{itemize}
 $\Leftrightarrow$ \begin{itemize}
                                                                                  \item ($HW\ge s$) or ($HW<s$ and $hw<\frac{Ss}{s-HW}$)
                                                                                  \item ($HW<s$ and $hw\ge\frac{Ss}{s-HW}$) and $rank(\Mb)<min(hws,Ss+hwHW)$.
                                                                                \end{itemize}
\end{proof}

\noindent \textbf{Theorem}~\ref{thm:MRA theorem}.  if the matrix equations in \textbf{Assumption}~\ref{as:total equations} have a solution, then one solution format to the \textbf{Assumption}~\ref{as:total equations} is the following
\begin{eqnarray}
\Xb_{mra}=\Bb^{-}\Zb+(\Ib-\Bb^{-}\Bb)\Yb\Ab^{-},
\end{eqnarray}
According to \cite{ben2003generalized}, the general solution space is
\begin{eqnarray}
\Xb=\Xb_{mra}+(\Ib-\Bb^{-}\Bb)\Wb(\Ib-\Ab\Ab^{-}),
\end{eqnarray}
where $\Wb\in R^{HW\times S}$ is an arbitrary matrix.
\begin{proof}
"$\Rightarrow$" $\forall \Wb\in R^{HW\times S},\Ab^{-}\in\Ab\{-\},\Bb^{-}\in\Bb\{-\}$, according to Theorem~\ref{thm:solution exsits coditions},
\begin{eqnarray}
\Bb\Yb = \Zb\Ab\\
\Yb=\Yb\Ab^{-}\Ab\\
\Zb=\Bb\Bb^{-}\Zb,
\end{eqnarray}
we have
\begin{eqnarray}
\Bb\Xb&=&\Bb(\Xb_{mra}+(\Ib-\Bb^{-}\Bb)\Wb(\Ib-\Ab\Ab^{-}))\nonumber\\
&=&\Bb\Xb_{mra}\nonumber\\
&=&\Bb(\Bb^{-}\Zb+(\Ib-\Bb^{-}\Bb)\Yb\Ab^{-})\nonumber\\
&=&\Bb\Bb^{-}\Zb\nonumber\\
&=&\Zb,
\end{eqnarray}
\begin{eqnarray}
\Xb\Ab&=&(\Xb_{mra}+(\Ib-\Bb^{-}\Bb)\Wb(\Ib-\Ab\Ab^{-}))\Ab\nonumber\\
&=&\Xb_{mra}\Ab\nonumber\\
&=&(\Bb^{-}\Zb+(\Ib-\Bb^{-}\Bb)\Yb\Ab^{-})\Ab\nonumber\\
&=&\Bb^{-}\Zb\Ab + \Yb\Ab^{-}\Ab - \Bb^{-}\Bb\Yb\Ab^{-}\Ab\nonumber\\
&=&\Bb^{-}\Zb\Ab + \Yb\Ab^{-}\Ab - \Bb^{-}\Zb\Ab\Ab^{-}\Ab\nonumber\\
&=&\Yb\Ab^{-}\Ab\nonumber\\
&=&\Yb.
\end{eqnarray}
"$\Leftarrow$"$\forall \Xb$, take $\Wb=\Xb-\Xb_{mra}$, it yields
\begin{eqnarray}
&&\Xb-(\Xb_{mra}+(\Ib-\Bb^{-}\Bb)(\Xb-\Xb_{mra})(\Ib-\Ab\Ab^{-}))\nonumber\\
&=&\Xb-\Xb_{mra}-(\Ib-\Bb^{-}\Bb)(\Xb-\Xb_{mra})(\Ib-\Ab\Ab^{-})\nonumber\\
&=& \Bb^{-}\Bb(\Xb-\Xb_{mra})+ (\Xb-\Xb_{mra})\Ab\Ab^{-} -\Bb^{-}\Bb(\Xb-\Xb_{mra})\Ab\Ab^{-}\nonumber\\
&=&\Bb^{-}(\Yb-\Yb)+(\Zb-\Zb)\Ab^{-}-\Bb^{-}(\Yb-\Yb)\Ab\Ab^{-}\nonumber\\
&=& 0.
\end{eqnarray}
\end{proof}

\noindent \textbf{Theorem}~\ref{thm:CS theorem}. If the matrix equations in \textbf{Assumption}~\ref{as:total equations} have a solution, then one solution format to the \textbf{Assumption}~\ref{as:total equations} is the following
\begin{eqnarray}
\Xb_{cs}=\Bb^{-}\Zb + (\Yb-\Bb^{-}\Zb\Ab) \Ab^{-},
\end{eqnarray}
and the general solution space is
\begin{eqnarray}
\Xb=\Xb_{cs}+(\Ib-\Bb^{-}\Bb)\Wb(\Ib-\Ab\Ab^{-}),
\end{eqnarray}
where $\Wb\in R^{HW\times S}$ is an arbitrary matrix.

\begin{proof}
"$\Rightarrow$" $\forall \Wb\in R^{HW\times S},\Ab^{-}\in\Ab\{-\},\Bb^{-}\in\Bb\{-\}$, since the matrix equations in \textbf{Assumption}~\ref{as:total equations} have a solution, according to Theorem~\ref{thm:solution exsits coditions},
\begin{eqnarray}
\Bb\Yb = \Zb\Ab\\
\Yb=\Yb\Ab^{-}\Ab\\
\Zb=\Bb\Bb^{-}\Zb,
\end{eqnarray}
we have
\begin{eqnarray}
\Xb_{cs}=\Bb^{-}\Zb + (\Yb-\Bb^{-}\Zb\Ab) \Ab^{-}=\Yb\Ab^{-} +\Bb^{-}\Zb(\Ib-\Ab\Ab^{-}).
\end{eqnarray}
It yields
\begin{eqnarray}
\Xb\Ab&=&(\Xb_{cs}+(\Ib-\Bb^{-}\Bb)\Wb(\Ib-\Ab\Ab^{-}))\Ab\nonumber\\
&=&(\Yb\Ab^{-} +\Bb^{-}\Zb(\Ib-\Ab\Ab^{-}))\Ab^{-}\nonumber\\
&=&\Yb\Ab^{-}\Ab\nonumber\\
&=&\Yb,
\end{eqnarray}
\begin{eqnarray}
\Bb\Xb&=&\Bb(\Xb_{cs}+(\Ib-\Bb^{-}\Bb)\Wb(\Ib-\Ab\Ab^{-}))\nonumber\\
&=&\Bb(\Yb\Ab^{-} +\Bb^{-}\Zb(\Ib-\Ab\Ab^{-}))\nonumber\\
&=&\Zb\Ab\Ab^{-} +\Bb\Bb^{-}\Zb(\Ib-\Ab\Ab^{-})\nonumber\\
&=&\Zb\Ab\Ab^{-} +\Zb(\Ib-\Ab\Ab^{-})\nonumber\\
&=&\Zb.
\end{eqnarray}
"$\Leftarrow$"
$\forall \Xb$ satisfying \textbf{Assumption}~\ref{as:total equations}, take $\Wb=\Xb-\Xb_{cs}$, it yields
\begin{eqnarray}
&&\Xb-(\Xb_{cs}+(\Ib-\Bb^{-}\Bb)(\Xb-\Xb_{cs})(\Ib-\Ab\Ab^{-}))\nonumber\\
&=&\Xb-\Xb_{cs}-(\Ib-\Bb^{-}\Bb)(\Xb-\Xb_{cs})(\Ib-\Ab\Ab^{-})\nonumber\\
&=& \Bb^{-}\Bb(\Xb-\Xb_{cs})+ (\Xb-\Xb_{cs})\Ab\Ab^{-} -\Bb^{-}\Bb(\Xb-\Xb_{cs})\Ab\Ab^{-}\nonumber\\
&=&\Bb^{-}(\Yb-\Yb)+(\Zb-\Zb)\Ab^{-}-\Bb^{-}(\Yb-\Yb)\Ab\Ab^{-}\nonumber\\
&=& 0.
\end{eqnarray}
\end{proof}

\noindent\textbf{Theorem}~\ref{thm:CS=MRA theorem} If the matrix equation in \textbf{Assumption}~\ref{as:total equations} has a solution, for fixed $\Ab^{-}\in \Ab\{-\}$ and $\Bb^{-}\in \Bb\{-\}$ then $\Xb_{cs}=\Xb_{mra}$.

\begin{proof}
Since $\Zb\Ab=\Bb\Yb$,
\begin{eqnarray}
\Xb_{cs}=\Bb^{-}\Zb + (\Yb-\Bb^{-}\Zb\Ab) \Ab^{-}=\Bb^{-}\Zb + (\Yb-\Bb^{-}\Bb\Yb) \Ab^{-}=\Xb_{mra}.
\end{eqnarray}
\end{proof}

\noindent\textbf{Theorem}~\ref{thm:GSA} The Gram Schmidt adaptive method can be described as follows: Suppose $var(\Zb\Ab)$ is invertible. Define
\begin{eqnarray}
\Wb =var(\Zb\Ab)^{-1} cov(\Zb\Ab,\Zb)\in \Ab\{-\}
\end{eqnarray}
where
\begin{eqnarray}
cov(\Zb\Ab,\Zb)&=&\frac{1}{hw}(\Zb\Ab-\oneb\frac{\oneb^T}{hw}\Zb\Ab)^T(\Zb-\oneb\frac{\oneb^T}{hw}\Zb),
\end{eqnarray}
\begin{eqnarray}
var(\Zb\Ab) = \frac{1}{hw}(\Zb\Ab-\oneb\frac{\oneb^T}{hw}\Zb\Ab)^T(\Zb\Ab-\oneb\frac{\oneb^T}{hw}\Zb\Ab),
\end{eqnarray}
It yields
\begin{eqnarray}\label{eq:gsa equation}
\Xb_{gsa}=\Bb^{-}\Zb + (\Yb-\Bb^{-}\Zb\Ab) \Wb.
\end{eqnarray}

The Gram Schmidt adaptive method can be viewed as a special case of the representation in the Theorem~\ref{thm:CS theorem}. It yields
\begin{eqnarray}
\Yb = \Xb_{gsa} \Ab \\
\Zb = \Bb \Xb_{gsa}.
\end{eqnarray}

\begin{proof}
According to the format of Equation~\ref{eq:gsa equation}, we only need to show that $W$ is the generalized inverse of A that is $AWA= A$.
Here, we only need to show that $WA=1$.
\begin{eqnarray}
cov(\Zb\Ab,\Zb)\Ab&=&\frac{1}{hw}(\Zb\Ab-\oneb\frac{\oneb^T}{hw}\Zb\Ab)^T(\Zb-\oneb\frac{\oneb^T}{hw}\Zb)\Ab = var(\Zb\Ab),
\end{eqnarray}
it yields
\begin{eqnarray}
\Wb\Ab=var(\Zb\Ab)^{-1}var(\Zb\Ab)=\Ib.
\end{eqnarray}
Therefore, $X_{gsa}$ satisfies
\begin{eqnarray}
\Yb = \Xb_{gsa} \Ab \\
\Zb = \Bb \Xb_{gsa}.
\end{eqnarray}

\end{proof}

\noindent\textbf{Theorem}~\ref{thm:error XA for cs} If $\Xb_{recover}=\Vb\Zb+(\Yb-\Vb\Zb\Ab)\Wb$, then $\Xb_{recover}\Ab -\Yb = (\Ib-\Wb\Ab)(\Vb\Zb\Ab - \Yb)$.

\begin{proof}
\begin{eqnarray}
\Xb_{recover}\Ab -\Yb &=&  (\Vb\Zb\Ab+(\Yb-\Vb\Zb\Ab)\Wb\Ab)-\Yb\nonumber\\
&=&(\Vb\Zb\Ab+\Yb\Wb\Ab-\Vb\Zb\Ab\Wb\Ab)-\Yb\nonumber\\
&=&(\Vb\Zb\Ab - \Yb)(\Ib-\Wb\Ab).
\end{eqnarray}
\end{proof}

\noindent\textbf{Theorem}~\ref{thm:error BX for cs} If $\Xb_{recover}=\Vb\Zb\Ab+(\Yb-\Vb\Zb\Ab)\Wb$, then $\Bb\Xb_{recover} -\Zb = \Bb\Vb\Zb-\Zb+\Bb(\Yb-\Vb\Zb\Ab)\Wb$.

\begin{proof}
\begin{eqnarray}
\Bb\Xb_{recover} -\Zb &=&  \Bb(\Vb\Zb+(\Yb-\Vb\Zb\Ab)\Wb)-\Zb\nonumber\\
&=&\Bb\Vb\Zb+\Bb\Yb\Wb-\Bb\Vb\Zb\Ab\Wb-\Zb\nonumber\\
&=&\Bb\Vb\Zb-\Zb+\Bb(\Yb-\Vb\Zb\Ab)\Wb
\end{eqnarray}

\end{proof}

\noindent\textbf{Theorem}~\ref{thm:error XA for mra} If $\Xb_{recover}=\Vb\Zb\Ab+(\Yb-\Vb\Bb\Yb)\Wb$, then $\Xb_{recover}\Ab -\Yb = \Vb(\Zb\Ab-\Bb\Yb)-(\Ib-\Vb\Bb)\Yb(\Ib-\Wb\Ab)$.

\begin{proof}
\begin{eqnarray}
\Xb_{recover}\Ab -\Yb &=&  (\Vb\Zb\Ab+(\Yb-\Vb\Bb\Yb)\Wb\Ab)-\Yb\nonumber\\
&=&(\Vb\Bb\Yb+\Vb(\Zb\Ab-\Bb\Yb)+\Yb\Wb\Ab-\Yb-\Vb\Bb\Yb\Wb\Ab)\nonumber\\
&=&\Vb(\Zb\Ab-\Bb\Yb)-(\Ib-\Vb\Bb)\Yb(\Ib-\Wb\Ab).
\end{eqnarray}
\end{proof}

\noindent\textbf{Theorem}~\ref{thm:error BX for mra} If $\Xb_{recover}=\Vb\Zb\Ab+(\Yb-\Vb\Bb\Yb)\Wb$, then $\Bb\Xb_{recover} -\Zb = (\Bb\Vb-\Ib)\Zb+(\Bb-\Bb\Vb\Bb)\Yb\Wb$.

\begin{proof}
\begin{eqnarray}
\Bb\Xb_{recover} -\Zb &=&  \Bb(\Vb\Zb+(\Yb-\Vb\Bb\Yb)\Wb)-\Zb\nonumber\\
&=&\Bb\Vb\Zb+\Bb\Yb\Wb-\Bb\Vb\Bb\Yb\Wb-\Zb\nonumber\\
&=&\Bb\Vb\Zb-\Zb+\Bb\Yb\Wb-\Bb\Vb\Bb\Yb\Wb\nonumber\\
&=&(\Bb\Vb-\Ib)\Zb+(\Bb-\Bb\Vb\Bb)\Yb\Wb
\end{eqnarray}

\end{proof}

\noindent\textbf{Theorem}~\ref{thm: down sampling enhancement} With down-sampling enhancement(DSE), the prior component substitution method and prior multiresolution analysis methods are equivalent. That is 
If $\Bb=\Zb\Zb^{+}\hat{\Bb}$ and $\Ab=\Zb^{+}\hat{\Bb}\Yb$, then $\Xb_{mra}=\Bb^{-}\Zb+(\Ib-\Bb^{-}\Bb)\Yb\Wb=\Xb_{cs}=\Bb^{-}\Zb + (\Yb-\Bb^{-}\Zb\Ab) \Wb.$

\begin{proof}
We only need to prove $\Bb^{-}\Bb\Yb\Wb=\Bb\Zb\Ab\Wb$.
Since with down-sampling enhancement, $\Bb=\Zb\Zb^{+}\hat{\Bb}$ and $\Ab = \Zb^{+}\hat{\Bb}\Yb$, we have
$$\Bb^{-}\Bb\Yb\Wb=\Bb^{-}\Zb\Zb^{+}\hat{\Bb}\Yb\Wb=\Bb\Zb\Ab\Wb.$$
\end{proof}

\noindent\textbf{Theorem}~\ref{thm:total error} If the Assumption~\ref{as:total equations} has a solution, then the ground truth $\Xb$ and $\Xb_{mra}$ is different in the following,
\begin{eqnarray}
\Xb-\Xb_{mra}=(\Ib-\Bb^{-}\Bb)\Xb(\Ib-\Ab\Ab^{-}).
\end{eqnarray}

\begin{proof}
\begin{eqnarray}
\Xb-(\Bb^{-}\Zb+(\Yb-\Bb^{-}\Bb\Yb)\Ab^{-}&=&\Xb-(\Bb^{-}\Bb\Xb+(\Xb\Ab-\Bb^{-}\Bb\Xb\Ab)\Ab^{-}\nonumber\\
&=&(\Ib-\Bb^{-}\Bb)\Xb+(\Ib-\Bb\Bb^{-})\Xb\Ab\Ab^{-}\nonumber\\
&=&(\Ib-\Bb^{-}\Bb)\Xb(\Ib-\Ab\Ab^{-})
\end{eqnarray}
\end{proof}

\vfill

.






\subsection{Detailed dataset descriptions for Chikusei dataset and PAirMax dataset}
\textbf{Detailed dataset description for the Chikusei dataset.\cite{yokoya2016airborne}} 
This airborne hyperspectral dataset, acquired by a Headwall Hyperspec-VNIR-C sensor over Chikusei, Ibaraki, Japan, on July 29, 2014 (9:56-10:53 UTC+9), features 128 bands covering 363 nm to 1018 nm. The scene (center: 36.294946°N, 140.008380°E) covers agricultural and urban areas with dimensions of 2517x2335 pixels and a 2.5 m ground sampling distance. Field surveys and visual interpretation of high-resolution Canon EOS 5D Mark II images collected alongside the flight provided ground truth labels for 19 classes(not used). The dataset and ground truth are distributed in ENVI and MATLAB formats for scientific use.

\textbf{Detailed dataset description for the PAirMax dataset.\cite{vivone2021benchmarking}} 
The PAirMax dataset consists of 14 scenes selected as representative examples of the heterogeneity encountered in Pan-sharpening applications. The dataset includes 14 Full-Resolution (FR) datasets.  14 Reduced-Resolution (RR) datasets were generated from these FR test cases. Most images depict urban areas, presenting significant pansharpening challenges such as accurately rendering high-contrast features (e.g., building edges), sub-resolution details, and avoiding spectral smearing across adjacent regions with differing spectral properties. Some scenes feature natural land cover types, including vegetation (agricultural fields, meadows, forests), and water bodies. Vegetated areas pose challenges due to the distinct reflectance difference between visible and NIR wavelengths (caused by chlorophyll) and textured patterns with sub-pixel size, potentially leading to spectral and spatial distortions. Water regions are included to assess spectral reconstruction, particularly relevant for sensors with bathymetry-optimized bands (e.g., WorldView-2/3), and often feature sharp edges against urban embankments. The scenes were acquired across different seasons, introducing variations in illumination conditions, such as dimmer intensities and elongated shadows in winter imagery.  All PAN-MS bundles exhibit a spatial resolution ratio of 4:1 (PAN has 16 times more pixels than each MS band). MS bands cover the Visible and Near-Infrared (VNIR) domain, with either four bands (GeoEye-1, WorldView-2, SPOT-7) or eight bands (WorldView-2/3). The 14 images in PAirMax were derived from original PAN+MS bundles acquired under clear-sky conditions with negligible cloud cover (especially within the selected cropped areas).

\subsection{Synthetic PAN and HS images in synthetic experiments.}
\begin{figure*}[htbp]
  \centering
  \begin{minipage}{0.6\textwidth}
    \centering
    \begin{subfigure}{\linewidth}
    \centering
    \includegraphics{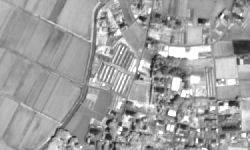}
    \caption{Generated pan}\label{fig:Generated Pan}
    \label{fig:left_image}
    \end{subfigure}
  \end{minipage}
  \hspace{0.1\textwidth}
  \begin{minipage}{0.25\textwidth}
    \centering
    \begin{subfigure}{\linewidth}
      \centering
      \includegraphics{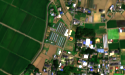}
      \caption{{RGB}}
      \label{fig:right_image1}
    \end{subfigure}
    \quad
    \begin{subfigure}{\linewidth}
      \centering
      \includegraphics{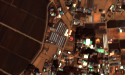}
      \caption{OAP}
      \label{fig:right_image2}
    \end{subfigure}
  \end{minipage}
  \caption{Synthetic PAN and HS images. RGB represents the red green blue channel and OAP represents the other selected three channels.}
  \label{fig:images_placement}
\end{figure*}

\subsection{Down-sampling and up-sampling parameters illustration}
In the comparative experiment, we use bilinear down-sampling and bilinear up-sampling. In the synthetic experiment, we use mean down-sampling to create the data and use bilinear down-sampling and bilinear up-sampling for calculation of our method. In ablation study and diffusion-related experiment, we use mean down-sampling and nearest neighborhood up-sampling. 

\subsection{Reduced-resolution comparative experiments}

In the reduced-resolution comparative experiments, our method does not perform outstandingly because it has not been refined by searching the general solution space with the diffusion prior.
\begin{table}[]
\centering
\caption{Reduced-resolution comparative experiments on WV3 New York Dataset}
\label{tab:Reduced resolution comparison experiments on WV3 New York Dataset}
\resizebox{0.48\textwidth}{!}{
\begin{tabular}{lccc}
\hline
\textbf{Method} & \textbf{Q2n}($\uparrow$)&\textbf{SAM}($\downarrow$)&\textbf{ERGAS}($\downarrow$)\\\hline
GT&1.0000&0.0000&0.0000\\\hline
EXP&0.6555&7.2344&8.2676\\\hline
BT-H&0.9261&6.4689&3.9948\\\hline
BDSD-PC&0.9334&6.8699&3.9157\\\hline
C-GSA&0.9235&6.7104&4.0751\\\hline
SR-D&0.9098&6.6782&4.4543\\\hline
MTF-GLP-HPM-R&0.9245&7.0266&4.0949\\\hline
MTF-GLP-FS&0.9246&6.7789&4.0691\\\hline
TV&0.9294&6.6165&4.0824\\\hline
PanNet&0.9217&6.9522&4.3506\\\hline
DRPNN&0.9224&7.4331&4.2887\\\hline
MSDCNN&0.9093&7.5705&4.4867\\\hline
BDPN&0.9198&7.7437&4.4823\\\hline
DiCNN&0.8558&8.1309&5.6302\\\hline
PNN&0.8830&12.6763&6.8061\\\hline
APNN&0.9119&7.6778&4.5471\\\hline
FusionNet&0.8534&8.4185&6.1608\\\hline
PCS/PMRA &0.8703&9.9481&5.6316\\\hline
\end{tabular}}
\end{table}
\end{document}